%% file: main.tex
\documentclass[journal]{IEEEtran}
\usepackage{cite}
\usepackage[switch,pagewise]{lineno}
\usepackage{url,subfigure}
\usepackage[hidelinks,hypertexnames=false]{hyperref}
\usepackage{booktabs}
\usepackage{multirow}
\usepackage{amssymb}
\usepackage{graphicx}
\usepackage{algorithm}
\usepackage{makecell}
\usepackage{ulem}
\usepackage{color}
\usepackage[table]{xcolor}
\usepackage{cuted}
\usepackage{capt-of}
\input{packages.tex}

\newcommand{\egi}{\textit{e.g.}}

\newcommand{\settablefont}{\fontsize{6.9}{11.8}\selectfont}

\begin{document}
\normalem
\title{Online, Target-Free LiDAR-Camera Extrinsic Calibration via Cross-Modal Mask Matching}
\author{
Zhiwei Huang$^{\orcidicon{0009-0008-7084-052X}\,}$,~\IEEEmembership{Graduate Student Member,~IEEE},
Yikang Zhang$^{\orcidicon{0009-0003-8840-392X}\,}$,~\IEEEmembership{Graduate Student Member,~IEEE},
\\
Qijun Chen$^{\orcidicon{0000-0001-5644-1188}\,},$~\IEEEmembership{Senior Member,~IEEE},
and Rui Fan$^{\orcidicon{0000-0003-2593-6596}\,}$,~\IEEEmembership{Senior Member,~IEEE}
\thanks{This research was supported by the National Science and Technology Major Project under Grant 2020AAA0108101, the National Natural Science Foundation of China under Grant 62233013, the Science and Technology Commission of Shanghai Municipal under Grant 22511104500, the Fundamental Research Funds for the Central Universities, and Xiaomi Young Talents Program. (\emph{Corresponding author: Rui Fan})}
\thanks{Zhiwei Huang, Yikang Zhang, Qijun Chen, and Rui Fan are with the Department of Control Science \& Engineering, the College of Electronics \& Information Engineering, Shanghai Institute of Intelligent Science and Technology, Shanghai Research Institute for Intelligent Autonomous Systems, the State Key Laboratory of Intelligent Autonomous Systems, and Frontiers Science Center for Intelligent Autonomous Systems, Tongji University, Shanghai 201804, China (e-mails: \{zhiweihuang, yikangzhang, qjchen\}@tongji.edu.cn, rui.fan@ieee.org).}
}
\maketitle	

\begin{strip}
\vspace{-9.0em}
\centering
\includegraphics[width=0.99999\textwidth]{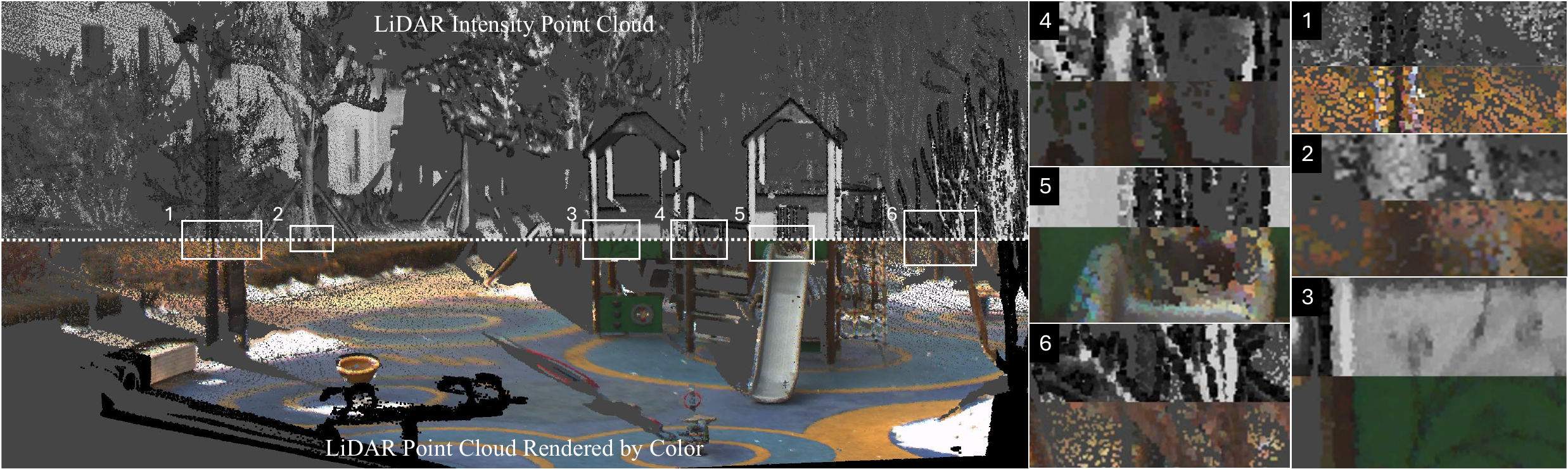}
\vspace{-1.0em}
\captionof{figure}{Visualization of the experimental results achieved using our proposed online, target-free LCEC algorithm.
}
\label{fig.cover}
\end{strip}

\begin{abstract}
LiDAR-camera extrinsic calibration (LCEC) is crucial for data fusion in intelligent vehicles. Offline, target-based approaches have long been the preferred choice in this field. However, they often demonstrate poor adaptability to real-world environments. This is largely because extrinsic parameters may change significantly due to moderate shocks or during extended operations in environments with vibrations. In contrast, online, target-free approaches provide greater adaptability yet typically lack robustness, primarily due to the challenges in cross-modal feature matching. Therefore, in this article, we unleash the full potential of large vision models (LVMs), which are emerging as a significant trend in the fields of computer vision and robotics, especially for embodied artificial intelligence, to achieve robust and accurate online, target-free LCEC across a variety of challenging scenarios. Our main contributions are threefold: we introduce a novel framework known as MIAS-LCEC, provide an open-source versatile calibration toolbox with an interactive visualization interface, and publish three real-world datasets captured from various indoor and outdoor environments. The cornerstone of our framework and toolbox is the cross-modal mask matching (C3M) algorithm, developed based on a state-of-the-art (SoTA) LVM and capable of generating sufficient and reliable matches. Extensive experiments conducted on these real-world datasets demonstrate the robustness of our approach and its superior performance compared to SoTA methods, particularly for the solid-state LiDARs with super-wide fields of view. Our toolbox and datasets are publicly available at mias.group/MIAS-LCEC.
\end{abstract}


\section{Introduction}
\label{sec.intro}

\subsection{Background}
\label{sec.intro_background}

\IEEEPARstart{L}{iDARs} provide accurate spatial geometric information, while cameras capture rich textural details \cite{arnold2019survey,fan2023autonomous,bai2022transfusion,ai2023lidar}. Fusing data from both sensors enables intelligent vehicles to achieve more comprehensive 3D environmental perception \cite{fan2020sne,cui2021deep,li2022deepfusion,li2024roadformer,zhang2023fs,ma2022computer}. LiDAR-camera extrinsic calibration (LCEC) forms the foundation for this data fusion process \cite{zhu2020online,wu2024s3mnet,sun2022atop,ou2023cross}, as illustrated in Fig. \ref{fig.cover}. It basically estimates an extrinsic matrix $^{C}_{L}\boldsymbol{T}$, defined as follows \cite{zhao2023dive}:
\begin{equation}
{^{C}_{L}\boldsymbol{T}} = 
\begin{pmatrix}
{^{C}_{L}\boldsymbol{R}} & {^{C}_{L}\boldsymbol{t}} \\
\boldsymbol{0}^\top & 1
\end{pmatrix}
\in{SE(3)},
\label{eq.lidar_to_camera_point}
\end{equation}
where $^{C}_{L}\boldsymbol{R} \in{SO(3)}$ represents the rotation matrix, $^{C}_{L}\boldsymbol{t}$ denotes the translation vector, and $\boldsymbol{0}$ represents a column vector of zeros. 
In this article, the symbols in the superscript and subscript denote the source and target sensors, respectively. When the camera intrinsic matrix $\boldsymbol{K}$ is known, a 3D LiDAR point $\boldsymbol{p}^{L}=(x^L;y^L;z^L)$ can be projected onto a 2D image pixel $\boldsymbol{{p}} = (u;v)$ using the following expression:
\begin{equation}
\tilde{\boldsymbol{p}} = \frac{\boldsymbol{K}({^{C}_{L}\boldsymbol{R}}\boldsymbol{p}^{L} + {^{C}_{L}\boldsymbol{t}})}{({^{C}_{L}\boldsymbol{R}}\boldsymbol{p}^{L} + {^{C}_{L}\boldsymbol{t}})^\top\boldsymbol{1}_{z}},
\end{equation}
where $\tilde{\boldsymbol{p}}$ represents the homogeneous coordinates of $\boldsymbol{{p}}$ and $\boldsymbol{1}_{z}=(0;0;1)$. While extensive research on offline, target-based LCEC has yielded numerous effective and robust algorithms over decades, online, target-free methods, especially for solid-state LiDARs, remain less explored \cite{tang2023robust}. Consequently, this study aims to bridge this gap by leveraging state-of-the-art (SoTA) large vision models (LVMs).

\subsection{Existing Challenges and Motivation}
\label{sec.intro_existing_challenges}

Existing online, target-free LCEC approaches first extract distinctive features, \egi, line/edge features \cite{lv2015automatic, wang2022temporal, yuan2021pixel, castorena2016autocalibration,tang2023robust,zhang2021line}, point features \cite{koide2023general,ye2021keypoint}, or semantic features \cite{wang2022automatic, ma2021crlf, li2018automatic, liao2023se,han2021auto}, from RGB images and LiDAR point clouds (spatial coordinates along with reflection intensities). These features are subsequently matched to produce cross-modal correspondences.

While line/edge feature-based LCEC approaches are highly efficient, their effectiveness is often limited by the requirement for sufficient and properly distributed edge features, confining their applicability to particular environments \cite{tang2023robust,zhang2021line}. First, edge detection algorithms typically identify straight lines, resulting in inadequate edge representations for objects with spherical or cylindrical geometries \cite{yuan2021pixel,wang2022temporal}. Moreover, in scenarios where edges are predominantly aligned in one direction, the constraints may not be sufficient to uniquely determine the extrinsic parameters \cite{lv2015automatic, castorena2016autocalibration}. Such scenarios can lead to solvers converging on local optima. Additionally, an uneven distribution of edges in an image can result in weak constraints that are susceptible to being influenced by measurement noise \cite{yuan2021pixel}.

On the other hand, point feature-based LCEC approaches require distinctive 2D image pixels and 3D LiDAR points, characterized by significant changes in intensity or depth across all dimensions. However, this dependency may lead to a scarcity of viable correspondences, especially in low-texture environments \cite{ye2021keypoint}. Additionally, these methods often overlook the alignment of the field of view (FoV) between LiDAR and camera, resulting in excess of irrelevant points within the LiDAR point clouds, adversely impacting the LCEC process's stability \cite{koide2023general}.

Recent advances in deep learning techniques have spurred extensive exploration of semantic feature-based LCEC approaches. Although these approaches have shown compelling performance in specific scenarios, such as parking lots, they predominantly rely on curated, pre-defined objects, \egi, vehicles \cite{li2018automatic}, lanes \cite{ma2021crlf}, poles \cite{wang2022automatic}, and stop signs \cite{han2021auto}. However, challenges such as domain shift (caused by differences in appearance, lighting conditions, or object distributions) and annotation inconsistency (different datasets often have diverse annotations) often impair the ability of these algorithms to generalize effectively across new, unseen scenarios.

Since 2023, LVMs have rapidly emerged as a focal point in the fields of computer vision and robotics. Models like Segmentation Anything (SAM) \cite{kirillov2023segment} and DINOv2 \cite{oquab2023dinov2} by Meta AI, have attracted significant attention and interest due to their exceptional generalizability across new, complex, and challenging scenarios \cite{liu2024playing}. Therefore, this study aims to leverage SoTA LVMs to extract more informative features and develop a more robust cross-modal feature matching strategy, thereby improving the overall performance of online, target-free LCEC.

\subsection{Novel Contributions}
\label{intro_novel_contributions}

Therefore, in this article, we move one step forward in the field of online, target-free LCEC by unleashing the potential of MobileSAM \cite{zhang2023faster}, a SoTA LVM for image segmentation. First, we develop an online, target-free LCEC approach, referred to as MIAS-LCEC, which employs a novel coarse-to-fine strategy to accurately estimate LiDAR-camera extrinsic parameters. To minimize the modality discrepancy, we formulate the 3D-2D feature matching as a 2D-2D feature matching problem by introducing a virtual camera (whose pose is iteratively updated) to project the given LiDAR point cloud, thereby generating a LiDAR intensity projection (LIP) image, which appears as if it were taken from the perspective of the actual camera. This addresses the oversight of FoV alignment in the prior study \cite{koide2023general} and helps achieve more effective and robust 3D-2D correspondence matching. Subsequently, both the LIP and RGB images undergo segmentation using MobileSAM. These segmentation results are then processed using a novel cross-modal mask matching (C3M) algorithm, capable of generating sparse yet reliable matches, which are propagated to target masks for dense matching. Finally, the obtained correspondences serve as inputs for a Perspective-n-Point (PnP) solver to derive the extrinsic matrix. Additionally, we launch a powerful toolbox with an interactive visualization interface. This toolbox also incorporates the manual calibration functionality, thereby further improving its utility. We collect three real-world datasets (from a variety of indoor and outdoor environments under various scenarios as well as different weather and illumination conditions) using a CMOS camera and different types of solid-state LiDARs to comprehensively evaluate the performance of LCEC algorithms. Through extensive experiments conducted on these datasets, our proposed MIAS-LCEC demonstrates superior robustness and accuracy compared to other SoTA online, target-free approaches. Moreover, it achieves a similar performance to that of an offline, target-based algorithm. Our toolbox and datasets are publicly available at \url{mias.group/MIAS-LCEC}.

In a nutshell, our main contributions are as follows:
\begin{itemize}
    \item {
    MIAS-LCEC, an online, target-free LCEC approach, which employs a novel coarse-to-fine strategy to accurately estimate LiDAR-camera extrinsic parameters by 
    unleashing the potential of MobileSAM, a SoTA LVM for image segmentation.
    }
    \item{
    C3M, a novel and robust cross-modal feature matching algorithm, capable of generating dense and reliable correspondences.
    }
    \item{
    A versatile LCEC toolbox with an interactive visualization interface and capable of conducting online, target-free calibration and manual calibration. 
    }
    \item {
    Three real-world datasets (containing dense 4D LiDAR point clouds and RGB images captured from a variety of indoor and outdoor environments), created to comprehensively evaluate the performance of LCEC algorithms.
    } 
\end{itemize}

\subsection{Article Structure}
\label{sec.outline}
The remainder of this article is structured as follows:
Sect. \ref{sec.related_work} reviews SoTA approaches in LCEC.
Sect. \ref{sec.methodology} introduces MIAS-LCEC, our proposed online, target-free LCEC algorithm.
Sect. \ref{sec.experiments} presents experimental results and compares our method with SoTA methods.
Finally, in Sect. \ref{sec.conclusion}, we conclude this article and discuss potential future research directions.

\section{Related Work}
\label{sec.related_work}
Existing LCEC approaches are primarily categorized as either target-based or target-free based on whether the algorithm requires pre-defined features from both RGB images and LiDAR point clouds. The following two subsections discuss these two types of algorithms in detail. 

\subsection{Target-Based Approaches}
\label{sec.rel_target-based}

The SoTA target-based LCEC approaches \cite{cui2020acsc, beltran2022automatic, koo2020analytic, xie2022a4lidartag, yan2023joint, tsai2021optimising} are typically offline, relying on customized calibration targets (typically checkerboards). These targets enable the automatic detection of correspondences, as they provide distinct, recognizable features that can be easily identified in both LiDAR and camera data. For example, the study \cite{cui2020acsc} performs extrinsic calibration by extracting corner points of a printed checkerboard from LiDAR point clouds and RGB images. It then optimizes the calibration result by formulating a RANSAC-based PnP problem, which minimizes the Euclidean distances between the corresponding corners. In the recent study \cite{yan2023joint}, both intrinsic and extrinsic parameters are accurately estimated using a specially designed calibration target, which incorporates a checkerboard pattern and four specifically placed holes. While these methods achieve high calibration accuracy, their reliance on customized targets and the need for additional setup render it impractical for scenarios where robots operate in dynamically changing environments. Consequently, we introduce a fully target-free approach that provides greater applicability and flexibility, eliminating the need for specialized calibration targets. This approach enables robots to rapidly obtain high-precision extrinsic parameters anytime and anywhere.

\subsection{Target-Free Approaches}
\label{sec.rel_target-free}

To improve the environmental adaptability of LCEC, previous studies \cite{scaramuzza2007extrinsic,dhall1705lidar,bileschi2009fully} have shifted from relying on specific targets to extracting informative visual features directly from the environment. In early attempts, researchers manually identified cross-modal correspondences and conducted LCEC using the PnP pose estimation algorithm \cite{scaramuzza2007extrinsic,dhall1705lidar}. Nevertheless, this manual LCEC process is tedious and prone to errors introduced by the manually selected correspondences \cite{tang2023robust}.

Afterwards, traditional line/edge feature-based automatic LCEC approaches emerged. In studies such as \cite{lv2015automatic, castorena2016autocalibration}, LiDAR point intensities are first projected into the camera perspective, thereby generating a virtual image, namely an LIP image. Edges are then extracted from both the LIP and RGB images. By matching these cross-modal edges, the relative pose between the two sensors can be determined. Similarly, research by \cite{pandey2012automatic, pandey2015automatic} optimizes extrinsic calibration by maximizing the mutual information (MI) between LIP and RGB images. To address occlusion issues in \cite{pandey2012automatic, pandey2015automatic}, the study of HKU-Mars \cite{yuan2021pixel} employs a voxelization method to detect and extract 3D lines from the point cloud, achieving high accuracy in scenarios with rich 3D line features. While effective in specific scenarios with abundant features, these traditional methods heavily rely on well-distributed line/edge features, which can compromise calibration robustness. Moreover, the use of low-level image processing algorithms, such as Gaussian blur and the Canny operator, can introduce errors in edge detection, potentially fragmenting global lines and thus reducing overall calibration accuracy.

Advances in deep learning techniques have driven significant exploration into enhancing traditional line/edge feature-based algorithms. In \cite{tang2023robust}, edge detection with Transformer (EDTER) \cite{pu2022edter}, a deep neural network, is utilized to improve the accuracy of 2D edge detection in RGB images. Additionally, a supervoxel-based 3D line detection method was designed to detect global line features from 3D point clouds, improving the line-based method proposed in \cite{yuan2021pixel}. Despite its impressive performance, this approach still heavily relies on specific scenarios with abundant properly-distributed lines, ultimately limiting its applicability. 

To overcome this limitation,  several end-to-end deep learning-based algorithms \cite{schneider2017regnet, iyer2018calibnet,lv2021lccnet} have been developed. RegNet \cite{schneider2017regnet}, the first convolutional neural network, developed specifically for extrinsic parameter estimation, is trained by minimizing a loss function representing the distance between predicted and ground-truth parameters. However, RegNet requires retraining when sensor intrinsic parameters change. CalibNet \cite{iyer2018calibnet} improves RegNet by maximizing geometric and photometric consistency between point clouds and images, thereby regressing extrinsic parameters implicitly. Another end-to-end network, LCCNet \cite{lv2021lccnet}, introduces a cross-attention module to measure the similarity between point clouds and images. While these methods have demonstrated effectiveness on large-scale datasets like KITTI \cite{geiger2012we}, which primarily focuses on urban driving scenarios, their performance has not been extensively validated on other types of real-world datasets. Furthermore, their dependence on pre-defined sensor configurations (both LiDAR and camera) poses implementation challenges.

Inspired by end-to-end keypoint detection and matching neural networks, a recent study \cite{koide2023general} introduced Direct Visual LiDAR Calibration (DVL), a novel point-based method that utilizes SuperGlue \cite{sarlin2020superglue} to establish direct 3D-2D correspondences between LiDAR and camera data. Additionally, this study refines the estimated extrinsic matrix through direct LiDAR-camera registration by minimizing the normalized information distance, a mutual information-based cross-modal distance measurement loss. However, an oversight in aligning the FoV between the LiDAR and camera leads to the presence of numerous redundant points of interest within the LiDAR point clouds, adversely affecting the overall stability of the LCEC process. Therefore, in this article, we improve the LiDAR point intensity projection strategy to generate LIP images that are more similar to the RGB images.

To further improve the robustness of cross-modal feature matching for LCEC, several studies \cite{li2018automatic, ma2021crlf, wang2022automatic, han2021auto, liao2023se, zhu2020online} have resorted to semantic segmentation techniques. For instance, in \cite{li2018automatic}, parking vehicles are first detected and then used to register point clouds with images. The study \cite{zhu2020online} maximizes the overlapping area of vehicles as represented in both point clouds and images.
Similarly, \cite{ma2021crlf} accomplishes LiDAR and camera registration by aligning road lanes and poles, while \cite{han2021auto} employs stop signs as calibration primitives and refines results over time using a Kalman filter. Moreover, \cite{liao2023se} investigates the consistency of segmented edges between point clouds and images, and \cite{wang2022automatic} introduces an automatic registration method based on pole matching. However, as discussed earlier, domain shift and annotation inconsistency issues often hinder these algorithms from effectively generalizing to unseen, new scenarios. In this work, we take a pioneering step by leveraging SoTA LVMs to extract more informative features from both LIP and RGB images. Furthermore, we develop a more robust cross-modal feature matching strategy, which makes the LCEC process fully target-free, overcoming the previous reliance on specific semantic targets. By integrating these advancements, we aim to enhance the accuracy and robustness of LCEC algorithms, enabling them to perform effectively in diverse and challenging real-world environments.

\section{Methodology}
\label{sec.methodology}

\subsection{Algorithm Overview}
\label{sec.algo_overview}

\begin{figure}[t!]
    \centering
    \includegraphics[width=0.48\textwidth]{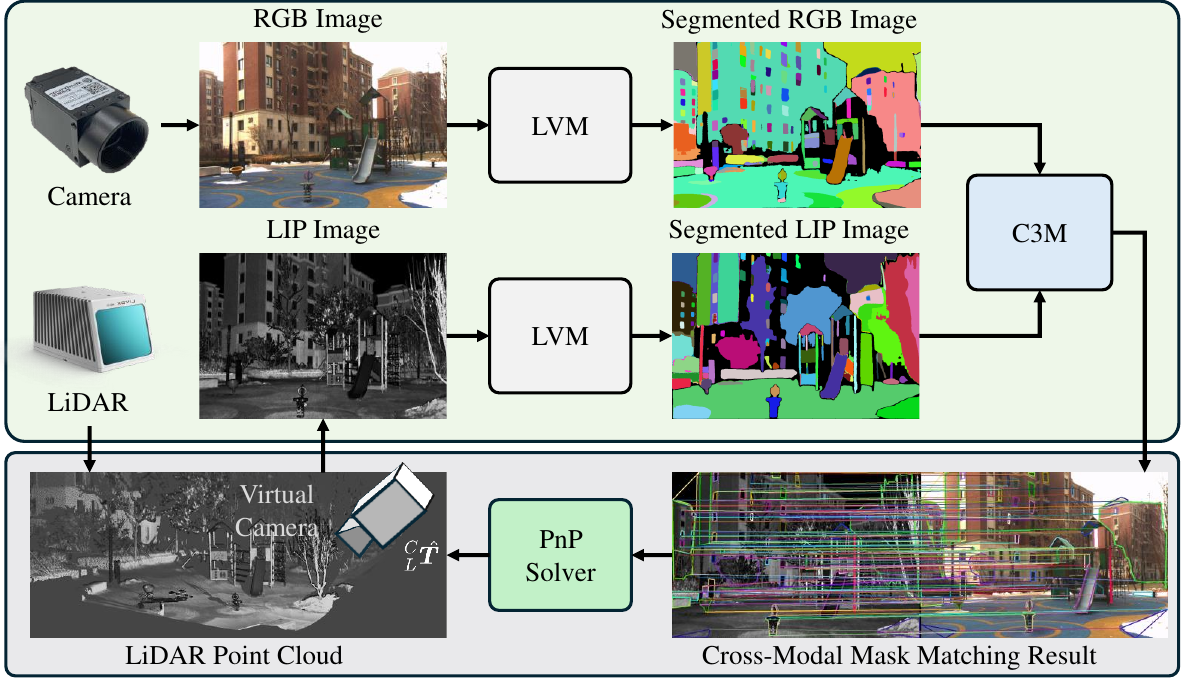}
    \caption{The pipeline of our proposed online, target-free LCEC algorithm.}
    \label{fig.algo_overview}
\end{figure}

Our proposed online, target-free LCEC algorithm MIAS-LCEC, as depicted in Fig. \ref{fig.algo_overview}, employs a novel coarse-to-fine pipeline. A virtual camera projects LiDAR point intensities into the camera perspective. Both the resulting LIP image and the RGB image are processed using MobileSAM, a SoTA LVM for image segmentation. Sufficient and reliable correspondences identified by our C3M strategy are then used as inputs for a PnP solver to estimate the extrinsic matrix ${^C_L}\boldsymbol{T}$.

Previous studies \cite{koide2023general,zhang2022multi} typically set up the virtual camera with a relative transformation ${^{V}_{L}\boldsymbol{T}}$ from LiDAR as follows:
\begin{equation}
{^{V}_{L}\boldsymbol{T}} = 
\begin{pmatrix}
\underbrace{
\begin{pmatrix}
0 & -1 & 0 \\
0 & 0 & -1 \\
1 & 0 & 0 \\
\end{pmatrix}}_{{^{V}_{L}\boldsymbol{R}}} & \underbrace
{
\begin{pmatrix}
0\\
0\\
0\\
\end{pmatrix}
}
_
{
{^{V}_{L}\boldsymbol{t}}
}
 \\
\boldsymbol{0}^\top & 1
\end{pmatrix}
\in{SE(3)},
\label{eq.other_transformation}
\end{equation}
thereby generating an LIP image $\boldsymbol{I}^L \in{\mathbb{R}^{H\times W \times 1}}$ to formulate the LCEC problem as a 2D feature matching problem, where $H$ and $W$ denote its height and width, respectively. Considering the image distortion introduced by different perspective views, (\ref{eq.other_transformation}) constrains the sensor setup to a captious relative transformation. Therefore, we propose a novel strategy to iteratively refine the virtual camera pose until the LIP image resembles one taken from the actual camera's perspective view. This iterative process can be expressed as follows:
\begin{equation}
\label{eq.virtual_interative_updates}
\lim_{k \to +\infty}{^{C}_{V}\boldsymbol{T}_k}=\lim_{k \to +\infty}({^{V}_{L}\boldsymbol{T}_k})^{-1}{^{C}_{L}\boldsymbol{T}}\approx 
\begin{pmatrix}
\boldsymbol{I} & \boldsymbol{0} \\
\boldsymbol{0}^\top & 1
\end{pmatrix},
\end{equation}
where the subscript $k$ denotes the $k$-th iteration, ${^{V}_{L}\boldsymbol{T}_k}$ represents the transformation from LiDAR to the virtual camera, and $\boldsymbol{I}$ denotes the identity matrix. 

Using the LIP image captured in each iteration and our proposed C3M in Sect. \ref{sec.c3m}, we can generate two sets $\mathcal{P}_k = \{\boldsymbol{p}_{k,1}, \dots, \boldsymbol{p}_{k,N_k} \}$ and $\mathcal{P}_k^L = \{\boldsymbol{p}_{k,1}^L, \dots, \boldsymbol{p}_{k,N_k}^L\} $, which store 2D pixels in the RGB image captured by camera and the corresponding 3D LiDAR points, respectively. 
The extrinsic matrix ${^{C}_{L}\hat{\boldsymbol{T}}_k}$ can then be effectively computed by minimizing the mean reprojection error as follows:
\begin{equation}
{^{C}_{L}\hat{\boldsymbol{T}}_k} = \underset{^C_L{\boldsymbol{T}_{k,i}}}{\arg\min} 
\underbrace {\frac{1}{N_k} \sum_{n=1}^{N_k}{\left\|\boldsymbol{K}(
{^{C}_{L}\boldsymbol{R}_{k,i}}
 \boldsymbol{p}_{k,n}^L + 
{^{C}_{L}\boldsymbol{t}_{k,i}}
) - {\boldsymbol{p}}_{k,n}\right\|_2}}_{\epsilon_k}, 
\label{eq.TmatEstimate}
\end{equation}
where ${^{C}_{L}\boldsymbol{T}_{k,i}}=\begin{pmatrix}
{^{C}_{L}\boldsymbol{R}_{k,i}} & {^{C}_{L}\boldsymbol{t}_{k,i}} \\
\boldsymbol{0}^\top & 1
\end{pmatrix}
\in{SE(3)}$ denotes the $i$-th PnP solution obtained using a selected subset of correspondences from $\mathcal{P}_k$ and $\mathcal{P}_k^L$, and $\epsilon_k$ represents the mean reprojection error with respect to ${^{C}_{L}\boldsymbol{T}_{k,i}}$.

Our MIAS-LCEC algorithm updates ${^{V}_{L}\boldsymbol{T}_{k+1}}$ with ${^{C}_{L}\hat{\boldsymbol{T}}_k}$. According to (\ref{eq.virtual_interative_updates}), ${^{V}_{L}\boldsymbol{T}_{k+1}}={^{C}_{L}\hat{\boldsymbol{T}}_k}\approx{^{C}_{L}\boldsymbol{T}}$ as the iterative process converges, minimizing the calibration error to the greatest extent. In practical applications, to optimize the trade-off between accuracy and efficiency, we terminate the iterative process when $\epsilon_{k+1} > \epsilon_{k}$, and select ${^{C}_{L}\hat{\boldsymbol{T}}_k}$ from the $k$-th iteration as the final calibration result, namely $^{C}_{L}{\boldsymbol{T}^{*}}$.

\subsection{Cross-Modal Mask Matching}
\label{sec.c3m}
In this article, we adopt a two-stage strategy to realize cross-modal mask matching, as detailed in Algorithm \ref{Algorithm.c3M}. Each stage consists of sequential coarse instance matching and fine-grained corner point matching. The first stage yields reliable yet sparse matches, from which we derive the parameters for an affine transformation to update the masks within the LIP image. In the second stage, we achieve dense mask matching by propagating the obtained reliable reference matches to the target masks. These dense matches are finally utilized as inputs for the PnP solver to obtain ${^{C}_{L}{\boldsymbol{T}}}$.

\begin{figure}
    \centering
    \includegraphics[width=0.48\textwidth]{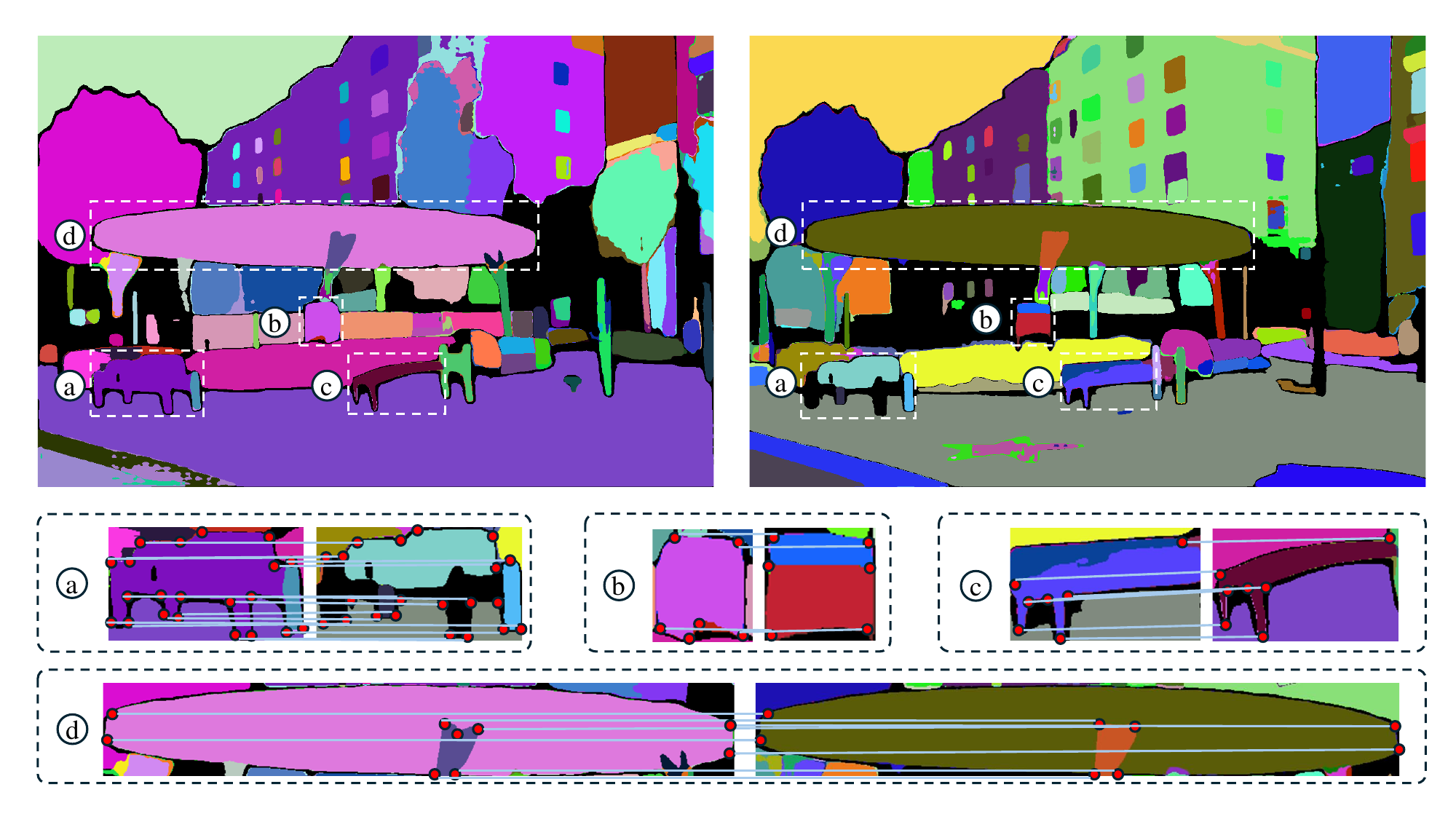}
    \caption{
    An example of two-stage, coarse-to-fine cross-modal mask matching result: (a)-(d) illustrate four examples of instance matching and corner point matching results. Potential errors produced by the LVM are greatly minimized through our strict match selection criterion.
    }
    \label{fig.C3MReduceError}
\end{figure}

\begin{figure}
    \centering
    \includegraphics[width=0.48\textwidth]{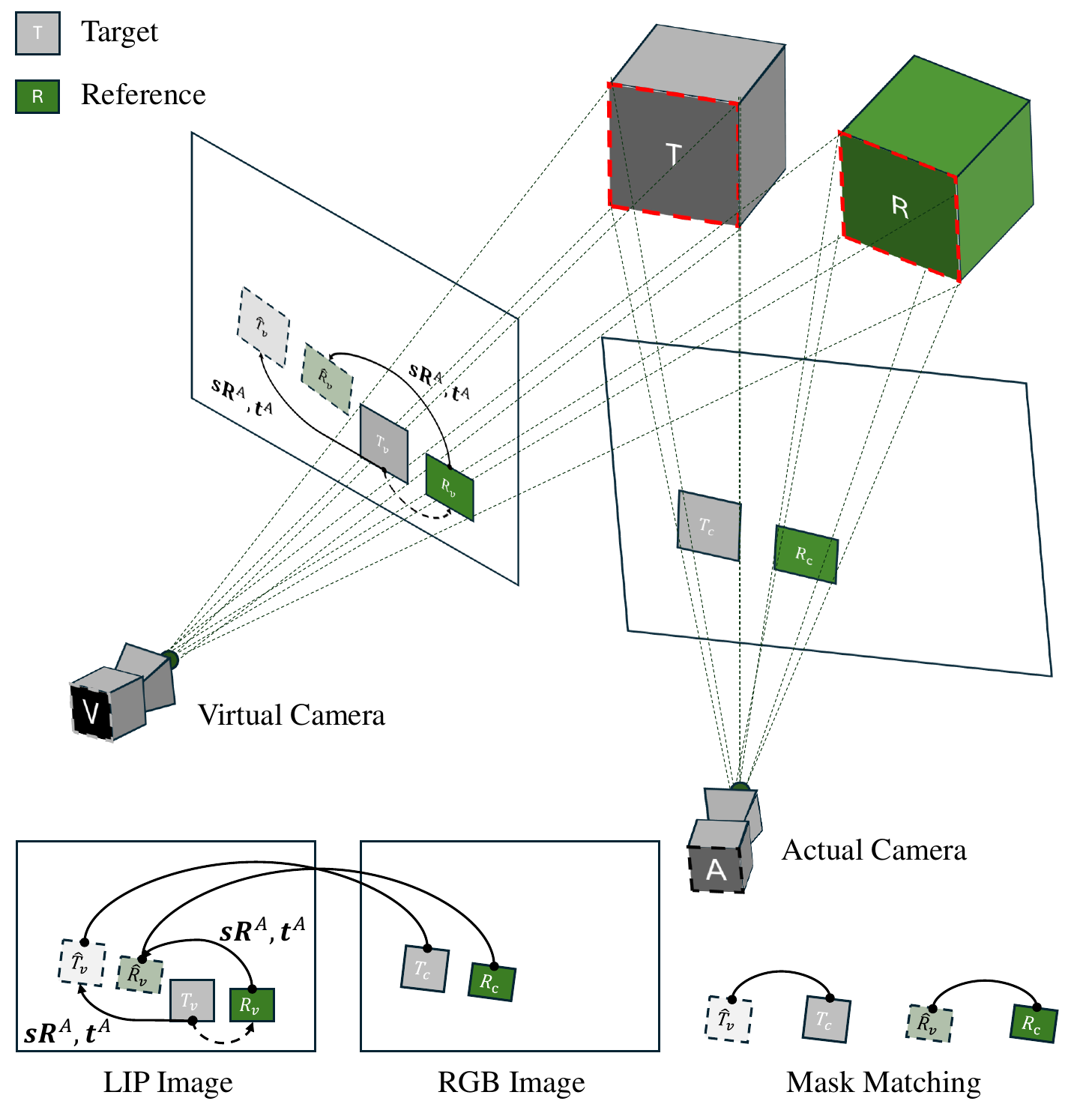}
    \caption{Upon matching the target masks $T_v$ and $T_c$, an affine transformation estimated from a pair of reference masks $R_v$ and $R_c$, is used to update the mask in the LIP image, so as to more accurately reflect the actual matching relationship.
}
\label{fig.method_affine_trans}
\end{figure}

As shown in Fig. \ref{fig.C3MReduceError}, our developed C3M strategy significantly minimizes potential matching errors produced by the LVM, primarily due to the strict match selection criterion. The corner points along the contours of masks detected within the LIP image and the RGB image using MobileSAM are represented by two sets: $\mathcal{C}^V = \{\boldsymbol{c}^V_1, \dots, \boldsymbol{c}^V_m\}$ and $\mathcal{C}^C = \{\boldsymbol{c}^C_1, \dots, \boldsymbol{c}^C_m\}$, respectively. An instance (bounding box), utilized to precisely fit around each mask, is centrally positioned at $\boldsymbol{o}^{V,C}$ and has a dimension of $h^{V,C}\times w^{V,C}$ pixels. To determine optimum instance matches, we construct a cost matrix $\boldsymbol{M}^I$, where the element at $\boldsymbol{x}=(i;j)$, namely:
\begin{equation}
\begin{aligned}                     
\boldsymbol{M}^I(\boldsymbol{x})=&\frac{1}{4}\Bigg(\frac{\left |w^C-w^V\right |}{w^C+w^V}+\frac{\left |h^C-h^V\right |}{h^C+h^V}+
\\
&2\bigg(1-\exp({-\frac{\left\|\hat{\boldsymbol{o}}^V-{\boldsymbol{o}}^C\right\|_2}{h^C+h^V+w^C+w^V}})\bigg)\Bigg)\in[0,1],
\end{aligned}
\label{eq.costfunc_mask}
\end{equation}
denotes the matching cost between the $i$-th instance from the LIP image and the $j$-th instance from the RGB image. $\hat{\boldsymbol{o}}^V$ is initially set as ${\boldsymbol{o}}^V$ during the sparse matching phase and subsequently updated using the above-mentioned affine transformation prior to the dense matching phase, as illustrated in Fig. \ref{fig.method_affine_trans}, so as to minimize the discrepancies arising from the differing perspectives between LiDAR and camera. A strict criterion is applied to achieve sparse yet reliable matching. Matches with the lowest costs in both horizontal and vertical directions are determined as the optimum coarse instance matches.

Subsequently, we determine corner point correspondences within the matched instances. Similarly, a cost matrix $\boldsymbol{M}^C$ is constructed, where the element at $\boldsymbol{y} = (r;s)$, namely:
\begin{equation}
\boldsymbol{M}^C(\boldsymbol{y}) = \frac{\|( \hat{\boldsymbol{c}}^V - \hat{\boldsymbol{o}}^V ) - (\boldsymbol{c}^C - \boldsymbol{o}^C )\|_2}{\|(\hat{\boldsymbol{c}}^V - \hat{\boldsymbol{o}}^V)\|_2 + \| (\boldsymbol{c}^C - \boldsymbol{o}^C) \|_2}\in[0,1],
\label{eq.costfunc_point}
\end{equation}
denotes the matching cost between the $r$-th corner point of a mask in the LIP image and the $s$-th corner point of a mask in the RGB image. $\hat{\boldsymbol{c}}^V$ is initialized as ${\boldsymbol{c}}^V$ during the sparse matching phase and updated using the same affine transformation prior to the dense matching phase. Correspondences with the lowest costs both horizontally and vertically are also determined to be optimum corner point matching results. Nevertheless, the first stage is considerably critical and cannot often provide the PnP solver with sufficient inputs. 

\begin{algorithm}[t!]
\caption{Cross-Modal Mask Matching}
\textbf{Require:}\\
Cross-modal masks, obtained from the LIP and RGB images.\\
\textbf{Stage 1 (Reliable sparse matching):} \\
(1) Construct the instance matching cost matrix $\boldsymbol{M}^I$ using (\ref{eq.costfunc_mask}).\\
(2) Select matched instances with low costs from $\boldsymbol{M}^I$ as reliable matches.\\
(3) Construct $\boldsymbol{M}^C$ using (\ref{eq.costfunc_point}) and match corner points. \\
(4) Estimate $s\boldsymbol{R}^A$ and $\boldsymbol{t}^A$ using (\ref{eq.affine_r})-(\ref{eq.affine_t}).\\
\textbf{Stage 2 (Dense mask matching):} \\
(1) Update all masks in the LIP image using $s\boldsymbol{R}^A$ and $\boldsymbol{t}^A$.\\
(2) Update $\boldsymbol{M}^I$ with the updated masks to obtain dense instance matching results. \\
(3) For each pair of matched instances, update their $\boldsymbol{M}^C$ using (\ref{eq.costfunc_point}) to determine corner point correspondences.\\
(4) Aggregate all corner point correspondences to form the sets $\mathcal{P}$ and $\mathcal{P}^L$.
\label{Algorithm.c3M}
\end{algorithm} 

Therefore, we apply an affine transformation to the masks within the LIP image to adjust $\boldsymbol{o}^V$ and $\boldsymbol{c}^V$, as follows:
\begin{equation}
\begin{cases}
\hat{\boldsymbol{c}}^V = s\boldsymbol{R}^A \boldsymbol{c}^V + \boldsymbol{t}^A\\
\hat{\boldsymbol{o}}^V = s\boldsymbol{R}^A \boldsymbol{o}^V + \boldsymbol{t}^A
\end{cases},
\label{eq.correction}
\end{equation}
where $\boldsymbol{R}^A\in{SO(2)}$ represents the rotation matrix, $\boldsymbol{t}^A$ denotes the translation vector, and $s$ represents the scaling factor. Given the critical nature of our designed sparse matching strategy, we can assume that after the affine transformation, any points within a given mask in the LIP image perfectly align with the corresponding points from the RGB image, and thus, $\hat{\boldsymbol{c}}^V = \boldsymbol{c}^C$ and $\hat{\boldsymbol{o}}^V = \boldsymbol{o}^C$. In this case, $\boldsymbol{R}^A$ can be obtained using the following expression:
\begin{equation}
\label{eq.affine_r}
\boldsymbol{R}^A = 
\begin{pmatrix}
\cos{\theta} & -\sin{\theta} \\
\sin{\theta} & \cos{\theta}
\end{pmatrix},
\end{equation}
where 
\begin{equation}
\begin{aligned}
{\theta} = \frac{1}{N}\sum_{i=1}^{N}{\bigg(\arctan{\frac{\boldsymbol{1}_y^\top({{\boldsymbol{c}}}^V_i  - {{\boldsymbol{o}}^V})}{\boldsymbol{1}_x^\top{({\boldsymbol{c}}}^V_i - {{\boldsymbol{o}}^V})}}} - 
\arctan{\frac{\boldsymbol{1}_y^\top({{\boldsymbol{c}}}^C_i - {{\boldsymbol{o}}^C})}{\boldsymbol{1}_x^\top({{\boldsymbol{c}}}^C_i - {{\boldsymbol{o}}^C})}}\bigg)
\end{aligned}
\label{eq.affine_theta}
\end{equation}
is the angle between the vectors originating from the mask centers and pointing to their respective matched corner points. $s$ can then be expressed as follows:
\begin{equation}
s = \frac{w^Ch^C}{w^Vh^V},
\label{eq.affine_s}
\end{equation}
which represents the ratio between the areas of the bounding boxes associated with the RGB image and the LIP image. Finally, according to (\ref{eq.correction}), $\boldsymbol{t}^A$ can be obtained as follows:
\begin{equation}
\boldsymbol{t}^A = {{\boldsymbol{o}}^C} - s\boldsymbol{R}^A{{\boldsymbol{o}}^V}.
\label{eq.affine_t}
\end{equation}
The remainder of this subsection delves into the relationship between a given pair of matched corner points within the reference and target masks, demonstrating the feasibility and reasonableness of propagating sparse, reliable mask matches. 

Given two 3D LiDAR points $\boldsymbol{q}^V$ (reference) and $\boldsymbol{p}^V$ (target) in the virtual camera coordinate system, their correspondences, $\boldsymbol{q}^C$ and $\boldsymbol{p}^C$, in the actual camera coordinate system can be established through the following transformations:
\begin{equation}
\boldsymbol{q}^C = {^C_V}\boldsymbol{R}\boldsymbol{q}^V +  {^C_V}\boldsymbol{t},
\label{eq.trans_reference}
\end{equation}
\begin{equation}
\boldsymbol{p}^C = {^C_V}\boldsymbol{R}\boldsymbol{p}^V +  {^C_V}\boldsymbol{t}.
\label{eq.trans_target}
\end{equation}
$\tilde{\boldsymbol{q}}_{v,c}$ and $\tilde{\boldsymbol{p}}_{v,c}$, the homogeneous coordinates of the corresponding 2D pixels of $\boldsymbol{q}^{V,C}$ and ${\boldsymbol{p}}^{V,C}$ in LIP and RGB images, can be obtained as follows:
\begin{align}
\left\{
\begin{aligned}
\tilde{\boldsymbol{p}}_v = \frac{\boldsymbol{K}}{(\boldsymbol{p}^V)^\top\boldsymbol{1}_{z}}\boldsymbol{p}^V, \ \ \tilde{\boldsymbol{p}}_c = \frac{\boldsymbol{K}}{(\boldsymbol{p}^C)^\top\boldsymbol{1}_{z}}\boldsymbol{p}^C, \\
\tilde{\boldsymbol{q}}_v = \frac{\boldsymbol{K}}{(\boldsymbol{q}^V)^\top\boldsymbol{1}_{z}}\boldsymbol{q}^V, \ \ \tilde{\boldsymbol{q}}_c = \frac{\boldsymbol{K}}{(\boldsymbol{q}^C)^\top\boldsymbol{1}_{z}}\boldsymbol{q}^C.
\end{aligned}
\right.
\label{eq.pproj}
\end{align}
Plugging (\ref{eq.pproj}) into (\ref{eq.trans_reference}) results in the affine transformation from $\tilde{\boldsymbol{q}}_v$ to $\tilde{\boldsymbol{q}}_c$ as follows:
\begin{equation}
\tilde{\boldsymbol{q}}_c = \underbrace{\frac{(\boldsymbol{q}^V)^\top\boldsymbol{1}_z}{(\boldsymbol{q}^C)^\top\boldsymbol{1}_z}\boldsymbol{K}({^C_V}\boldsymbol{R})\boldsymbol{K}^{-1}}_{
\boldsymbol{A}
}\tilde{\boldsymbol{q}}_v +  \underbrace{\frac{\boldsymbol{K}}{(\boldsymbol{q}^C)^\top\boldsymbol{1}_z}{^C_V}\boldsymbol{t}}_{
\boldsymbol{b}
}.
\label{eq.reference_trans}
\end{equation}
where $\boldsymbol{A}$ and $\boldsymbol{b}$ represent an affine transformation from ${\boldsymbol{q}}_v$ to ${\boldsymbol{q}}_c$. Combining (\ref{eq.trans_target}) and (\ref{eq.trans_reference}) results in the following expression:
\begin{equation}
\boldsymbol{p}^C = \boldsymbol{q}^C + {^C_V}\boldsymbol{R}(\boldsymbol{p}^V - \boldsymbol{q}^V).
\label{eq.reference_law}
\end{equation}
Plugging (\ref{eq.pproj}) into (\ref{eq.reference_law}) results in the following expression:
\begin{equation}
\begin{aligned}
\tilde{\boldsymbol{p}}_c = &
\frac{(\boldsymbol{q}^C)^\top\boldsymbol{1}_{z}}{(\boldsymbol{p}^C)^\top\boldsymbol{1}_{z}}\tilde{\boldsymbol{q}}_c \\
 + & \frac{(\boldsymbol{p}^V)^\top\boldsymbol{1}_{z}}{(\boldsymbol{p}^C)^\top\boldsymbol{1}_{z}}\boldsymbol{K}({^{C}_{V}\boldsymbol{R}})\boldsymbol{K^{-1}}(\tilde{\boldsymbol{p}}_v - \frac{(\boldsymbol{q}^V)^\top\boldsymbol{1}_{z}}{(\boldsymbol{p}^V)^\top\boldsymbol{1}_{z}}\tilde{\boldsymbol{q}}_v) \\
= &\frac{(\boldsymbol{p}^V)^\top\boldsymbol{1}_{z}}{(\boldsymbol{p}^C)^\top\boldsymbol{1}_{z}}\frac{(\boldsymbol{q}^C)^\top\boldsymbol{1}_{z}}{(\boldsymbol{q}^V)^\top\boldsymbol{1}_{z}}\underbrace{\frac{(\boldsymbol{q}^V)^\top\boldsymbol{1}_{z}}{(\boldsymbol{q}^C)^\top\boldsymbol{1}_{z}}\boldsymbol{K}({^{C}_{V}\boldsymbol{R}})\boldsymbol{K^{-1}}}_
 {
\boldsymbol{A}
 }
 \tilde{\boldsymbol{p}}_v   \\
+ & \frac{(\boldsymbol{q}^C)^\top\boldsymbol{1}_{z}}{(\boldsymbol{p}^C)^\top\boldsymbol{1}_{z}}\underbrace{\bigg(\tilde{\boldsymbol{q}}_c - \frac{(\boldsymbol{q}^V)^\top\boldsymbol{1}_{z}}{(\boldsymbol{q}^C)^\top\boldsymbol{1}_{z}}\boldsymbol{K}({^{C}_{V}\boldsymbol{R}})\boldsymbol{K^{-1}}\tilde{\boldsymbol{q}_v}\bigg)}_
{
\boldsymbol{b}
}.
\end{aligned}
\label{eq.LscLaw_ini}
\end{equation}
When $\boldsymbol{p}^V$ and $\boldsymbol{q}^V$ are close in depth, namely $(\boldsymbol{p}^V)^\top\boldsymbol{1}_{z}\approx (\boldsymbol{p}^C)^\top\boldsymbol{1}_{z}\approx(\boldsymbol{q}^V)^\top\boldsymbol{1}_{z}\approx (\boldsymbol{q}^C)^\top\boldsymbol{1}_{z}$, (\ref{eq.LscLaw_ini}) can be rewritten as follows:
\begin{align}
\tilde{\boldsymbol{p}}_c \approx 
\boldsymbol{A}
\tilde{\boldsymbol{p}}_v + 
\boldsymbol{b}.
\label{eq.LscLaw}
\end{align}
(\ref{eq.reference_trans}) and (\ref{eq.LscLaw}) indicate that $\tilde{\boldsymbol{q}}_{v,c}$ and $\tilde{\boldsymbol{p}}_{v,c}$ can share the same affine transformation when $\boldsymbol{q}^{V,C}$ and ${\boldsymbol{p}}^{V,C}$ are close in depth. 
In practice, we use the following affine transformation:
\begin{equation}
\begin{pmatrix}
\tilde{\boldsymbol{p}}_c & \tilde{\boldsymbol{q}}_c \\
\end{pmatrix}
=
\begin{pmatrix}
s{\boldsymbol{R}}^A & \boldsymbol{t}^A \\
\boldsymbol{0}^\top &1 \\
\end{pmatrix}
\begin{pmatrix}
\tilde{\boldsymbol{p}}_v & \tilde{\boldsymbol{q}}_v \\
\end{pmatrix}.
\label{eq.LscLaw_final}
\end{equation}
Therefore, the affine transformation of the target mask can be approximated by $s{\boldsymbol{R}^A}$ and ${\boldsymbol{t}^A}$, which are derived from the reference masks that are spatially close to the target mask. 

In the first stage of the C3M process, reference masks are not yet identified. Therefore, $s\boldsymbol{R}^A$ and $\boldsymbol{t}^A$ are not considered when constructing $\boldsymbol{M}^I$ and $\boldsymbol{M}^C$, and are initialized as
$
{s\boldsymbol{R}^{A}} = 
\begin{pmatrix}
1 & 0 \\
0 & 1 
\end{pmatrix}
$
and 
$
\boldsymbol{t}^A = 
(0;0)
$. In the second stage, they are determined using (\ref{eq.affine_r})-(\ref{eq.affine_t}), based on reliable reference masks identified in the first stage. The entire C3M process is efficient because it primarily focuses on 2D affine transformation, rather than complex 3D point cloud registration.

\section{Experiment}
\label{sec.experiments}
\begin{figure}[t!]
    \centering
    \includegraphics[width=0.48\textwidth]{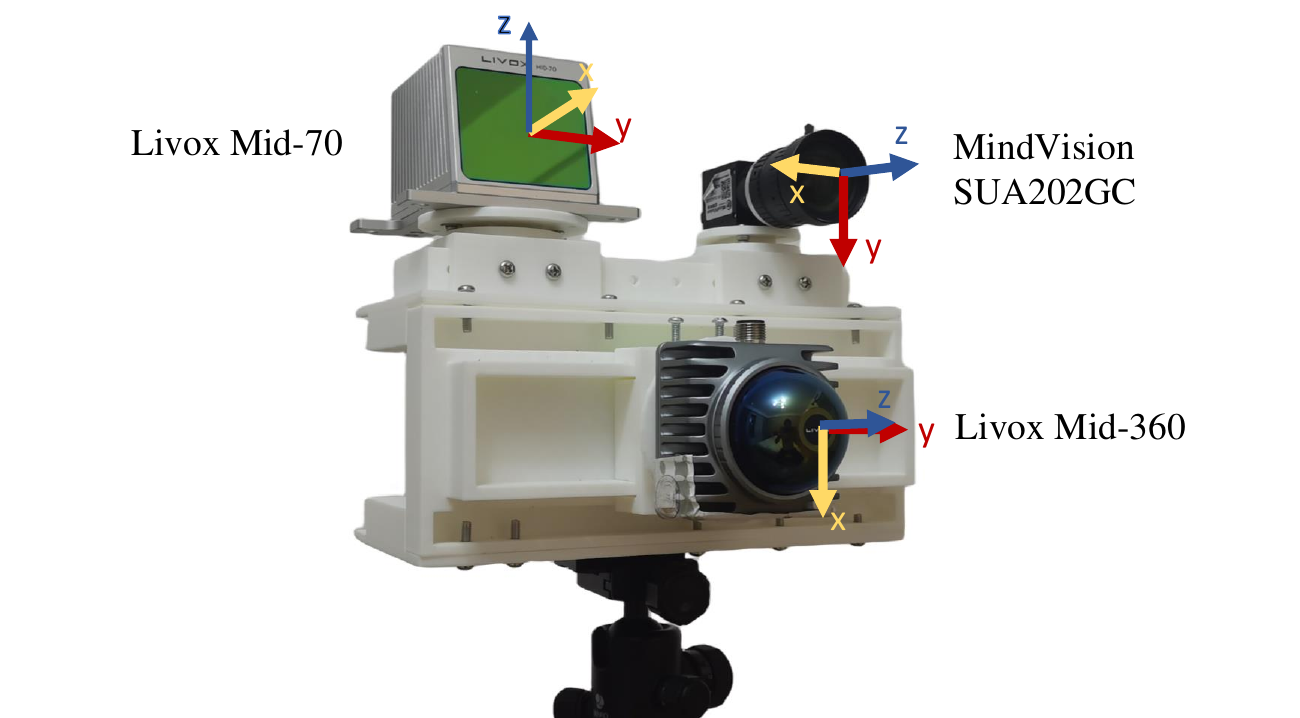}
    \caption{Our experimental setup, where two solid-state Livox LiDARs and one MindVision camera are utilized for data acquisition. }
    \label{fig.LcPlatform}
\end{figure}

\subsection{Experimental Setup}
\label{sec.exp_setup}
Our experimental setup, as shown in Fig. \ref{fig.LcPlatform}, consisting of two solid-state Livox LiDARs (Mid-70 and Mid-360) from DJI and an MV-SUA202GC global-shutter CMOS camera from MindVision, is used for cross-modal data collection. The Mid-70 and Mid-360 LiDARs both operate at a point rate of 200,000 points/s. However, the Mid-70 LiDAR captures dual returns, while the Mid-360 LiDAR captures only the first return. The RGB image resolution is 1,200$\times$800 pixels. The camera's intrinsic parameters are determined through offline calibration prior to the experiments and are presumed to remain constant.

We compare our method with four SoTA target-free LCEC approaches: CRLF \cite{ma2021crlf}, UMich \cite{pandey2015automatic}, HKU-Mars \cite{yuan2021pixel} and DVL \cite{koide2023general}, on two real-world datasets, MIAS-LCEC-TF70 and MIAS-LCEC-TF360. Additionally, to validate the effectiveness of our method in scenarios where targets are present, we also conduct comparisons with a classical offline target-based LCEC approach introduced in \cite{cui2020acsc} on the MIAS-LCEC-CB70 dataset.

Our algorithm was implemented on an Intel i7-14700K CPU and an NVIDIA RTX4070Ti Super GPU. The entire process, including data preprocessing, C3M, and extrinsic parameters optimization, takes approximately 15 to 70 seconds.

\begin{table*}[t!]
\caption{Quantitative comparisons with SoTA target-free LCEC approaches on the MIAS-LCEC-TF70 dataset. The best results are shown in bold type.}
\centering
\settablefont
\begin{tabular}{c|l|ccc|ccc|ccc|ccc}
\toprule
\multirow{3}*{Subsets}  &\multirow{3}*{Approach} &\multicolumn{6}{c|}{Rotation Error ($^\circ$)} &\multicolumn{6}{c}{Translation Error (m)} \\
\cline{3-14} 
{} & {} & \multirow{2}*{Yaw} & \multirow{2}*{Pitch} & \multirow{2}*{Roll}  & \multicolumn{3}{c|}{$e_r$ ($^\circ$)} &  \multirow{2}*{X} &  \multirow{2}*{Y} &  \multirow{2}*{Z}  &\multicolumn{3}{c}{$e_t$ (m)}\\
&&&&&Mean&Max&Min&&&&Mean&Max&Min\\
\hline
\hline
\multirow{5}{*}{\makecell{Residential Community}} 
& CRLF \cite{ma2021crlf}  &0.177	&1.581	&0.876	&1.594	&1.754	&1.537	&0.162	&0.429	&0.060	&0.464	&3.636	&0.136\\
& UMich \cite{pandey2015automatic}  &0.690	&1.502	&4.066	&4.829	&21.706	&0.214	&0.306	&0.151	&0.058	&0.387	&2.394	&0.072\\
& HKU-Mars \cite{yuan2021pixel} &1.704	&1.128	&1.113	&2.695	&7.036	&0.749	&0.829	&0.458	&0.397	&1.208	&4.405	&0.319\\
& DVL \cite{koide2023general} &0.107	&\textbf{0.089}	&\textbf{0.091}	&0.193	&\textbf{0.383}	&0.042	&0.034	&0.043	&\textbf{0.018}	&0.063	&\textbf{0.141}	&0.018\\
& \textbf{MIAS-LCEC (Ours)} &\textbf{0.049}	&0.138	&0.094	&\textbf{0.190}	&0.711	&\textbf{0.012}	&\textbf{0.032}	&\textbf{0.016}	&0.026	&\textbf{0.050}	&0.176	&\textbf{0.018}\\
\hline

\multirow{5}{*}{\makecell{Urban Freeway}}
& CRLF \cite{ma2021crlf}  &0.434	&1.584	&0.857	&1.582	&1.585	&1.581	&0.043	&0.130	&0.030	&0.140	&0.140	&0.140 \\
& UMich \cite{pandey2015automatic}  &0.471	&1.331	&1.464	&2.267	&6.531	&0.506	&\textbf{0.051}	&0.128	&0.063	&0.166	&0.241	&0.103 \\
& HKU-Mars \cite{yuan2021pixel}  &2.518	&1.610	&0.955	&2.399	&3.744	&0.337	&1.482	&1.125	&0.284	&1.956	&9.453	&0.148 \\
& DVL \cite{koide2023general}  &\textbf{0.212}	&0.140	&\textbf{0.045}	&0.298	&\textbf{0.420}	&0.090	&0.103	&0.056	&0.017	&0.124	&0.268	&0.064 \\
& \textbf{MIAS-LCEC (Ours)}  &0.216	&\textbf{0.111}	&0.083	&\textbf{0.291}	&0.514	&\textbf{0.062}	&0.070	&\textbf{0.052}	&\textbf{0.010}	&\textbf{0.111}	&\textbf{0.145	}&\textbf{0.047}
\\
\hline

\multirow{5}{*}{Building}
& CRLF \cite{ma2021crlf} &0.231	&1.409	&0.911	&1.499	&1.706	&1.465	&17.078	&8.692	&6.168	&20.165	&140.316	&0.140\\
& UMich \cite{pandey2015automatic} &0.927	&1.824	&15.097	&11.914	&22.778	&0.452	&0.544	&0.182	&0.407	&0.781	&1.497	&0.042\\
& HKU-Mars \cite{yuan2021pixel} &0.800	&1.391	&0.589	&1.814	&3.618	&0.118	&0.532	&0.262	&0.227	&0.706	&2.595	&0.059\\
& DVL \cite{koide2023general} &0.138	&\textbf{0.103}	&\textbf{0.049}	&0.200	&0.357	&0.059	&0.058	&0.047	&0.031	&0.087	&0.146	&0.052\\
& \textbf{MIAS-LCEC (Ours)} &\textbf{0.080}	&0.152	&0.074	&\textbf{0.198}	&\textbf{0.390}	&\textbf{0.051}	&\textbf{0.049}	&\textbf{0.036}	&\textbf{0.020}	&\textbf{0.072}	&\textbf{0.137}	&\textbf{0.024}\\
\hline

\multirow{5}{*}{Challenging Weather} 
& CRLF \cite{ma2021crlf}  &0.183	&1.647	&0.917	&1.646	&1.803	&1.552	&1.614	&0.867	&0.617	&2.055	&20.307	&0.137\\
& UMich \cite{pandey2015automatic}  &0.383	&1.635	&3.720	&1.851	&5.335	&0.294	&0.212	&0.109	&0.103	&0.310	&2.134	&0.043\\
& HKU-Mars \cite{yuan2021pixel} &1.110	&1.128	&1.840	&2.578	&11.647	&0.371	&0.772	&0.456	&0.469	&1.086	&4.934	&0.302\\
& DVL \cite{koide2023general} &\textbf{0.095}	&\textbf{0.100}	&0.069	&0.181	&\textbf{0.311}	&0.037	&0.029	&0.036	&\textbf{0.015}	&0.052	&\textbf{0.113}	&\textbf{0.017}\\
& \textbf{MIAS-LCEC (Ours)} &0.100	&0.112	&\textbf{0.059}	&\textbf{0.177}	&0.412	&\textbf{0.030}	&\textbf{0.027}	&\textbf{0.027}	&0.018	&\textbf{0.046}	&0.127	&0.021\\
\hline

\multirow{5}{*}{Indoor}
& CRLF \cite{ma2021crlf}  &0.164	&1.721	&1.010	&1.886	&2.201	&1.551	&27.895	&6.880	&6.368	&30.046	&90.926	&0.140\\
& UMich \cite{pandey2015automatic} &0.135	&1.531	&0.971	&2.029	&4.596	&0.448	&0.062	&0.062	&0.040	&0.109	&0.176	&0.020\\
& HKU-Mars \cite{yuan2021pixel} &1.065	&1.414	&1.572	&2.527	&5.458	&0.779	&0.165	&0.075	&0.122	&0.246	&0.893	&0.036\\
& DVL \cite{koide2023general} &0.261	&\textbf{0.194}	&\textbf{0.129}	&0.391	&0.879	&\textbf{0.193}	&0.022	&0.016	&\textbf{0.007}	&0.030	&0.078	&0.013\\
& \textbf{MIAS-LCEC (Ours)}  &\textbf{0.144}	&0.234	&0.175	&\textbf{0.363}	&\textbf{0.724}	&0.225	&\textbf{0.016}	&\textbf{0.011}	&0.011	&\textbf{0.024}	&\textbf{0.045}	&\textbf{0.012}\\
\hline

\multirow{5}{*}{\makecell{Challenging Illumination}} 
& CRLF \cite{ma2021crlf}  &0.088	&1.778	&1.156	&1.876	&2.141	&1.613	&18.080	&2.954	&3.886	&19.047	&34.444	&0.132 \\
& UMich \cite{pandey2015automatic} &0.361	&4.841	&2.924	&5.012	&12.087	&0.314	&0.108	&0.223	&0.148	&0.330	&0.679	&0.043 \\
& HKU-Mars \cite{yuan2021pixel} &7.762	&8.774	&7.235	&14.996	&38.750	&0.338	&1.609	&1.578	&1.679	&3.386	&10.520	&0.034 \\
& DVL \cite{koide2023general} &0.546	&1.573	&0.256	&1.747	&8.466	&0.207	&0.328	&0.144	&0.099	&0.377	&1.970	&\textbf{0.022} \\
& \textbf{MIAS-LCEC (Ours)}  &\textbf{0.347}	&\textbf{0.524}	&\textbf{0.180}	&\textbf{0.749}	&\textbf{1.554}	&\textbf{0.149}	&\textbf{0.087}	&\textbf{0.057}	&\textbf{0.030}	&\textbf{0.118}	&\textbf{0.220}	&0.033 \\
\hline
\hline
\multirow{5}{*}{All} 
& CRLF \cite{ma2021crlf} &0.197	&1.625	&0.946	&1.683	&2.201	&1.465	&10.051	&3.036	&2.589	&11.133	&140.316	&0.132 \\
& UMich \cite{pandey2015automatic} &0.485	&1.945	&4.272	&4.265	&22.778	&0.214	&0.217	&0.134	&0.115	&0.333	&2.394	&0.020 \\
&HKU-Mars \cite{yuan2021pixel} &2.140	&2.156	&1.988	&3.941	&38.750	&0.118	&0.806	&0.555	&0.475	&1.261	&10.520	&0.034 \\
&DVL \cite{koide2023general} &0.201	&0.292	&\textbf{0.104}	&0.423	&8.466	&0.037	&0.075	&0.050	&0.026	&0.100	&1.970	&0.013 \\
&\textbf{MIAS-LCEC (Ours)} &\textbf{0.133}	&\textbf{0.196}	&0.110	&\textbf{0.298}	&\textbf{1.554}	&\textbf{0.012}	&\textbf{0.040}	&\textbf{0.028}	&\textbf{0.019}	&\textbf{0.061}	&\textbf{0.220}	&\textbf{0.012} \\
\bottomrule
\end{tabular}
\label{tab.ResBySceneCmp}
\end{table*}

\subsection{Datasets}
\label{sec.datasets}
We have created the following three real-world datasets: \textbf{MIAS-LCEC-TF70} (target-free), \textbf{MIAS-LCEC-CB70} (target-based), and \textbf{MIAS-LCEC-TF360} (target-free), which are now publicly available for researchers to evaluate the performance of LCEC approaches: 
\begin{itemize}
\item MIAS-LCEC-TF70 is a diverse and challenging dataset that contains 60 pairs of 4D point clouds (including spatial coordinates with intensity data) and RGB images, collected using a Livox Mid-70 LiDAR and a MindVision SUA202GC camera, from a variety of indoor and outdoor environments, under various scenarios as well as different weather and illumination conditions. We divide this dataset into six subsets: residential community, urban freeway, building, challenging weather, indoor, and challenging illumination, to comprehensively evaluate the algorithm performance. 
\item MIAS-LCEC-CB70 contains 15 pairs of 4D point clouds and RGB images, all collected in our laboratory using a Livox Mid-70 LiDAR and a MindVision SUA202GC camera, with yaw angles ranging from -30$^\circ$ to +30$^\circ$, and the distances between the sensors and a calibration checkerboard pattern ranging from 3 m to 5 m. The checkerboard pattern comprises alternating white and black squares of equal size (8 cm $\times$ 8 cm).
\item MIAS-LCEC-TF360 contains 12 pairs of 4D point clouds and RGB images, collected using a Livox Mid-360 LiDAR and a MindVision SUA202GC camera from both indoor and outdoor environments. Since the Livox Mid-360 LiDAR has a scanning range of 360$^\circ$,  it produces a sparser point cloud compared to that generated by the Livox Mid-70 LiDAR. Additionally, the significant difference in the FoV between this type of LiDAR and the camera results in only a small overlap in the collected data. Consequently, this dataset is particularly well-suited for evaluating the adaptability of algorithms to challenging scenarios characterized by sparse point clouds and limited data overlap.
\end{itemize}

\begin{figure}[t!]
    \centering
    \includegraphics[width=0.48\textwidth]{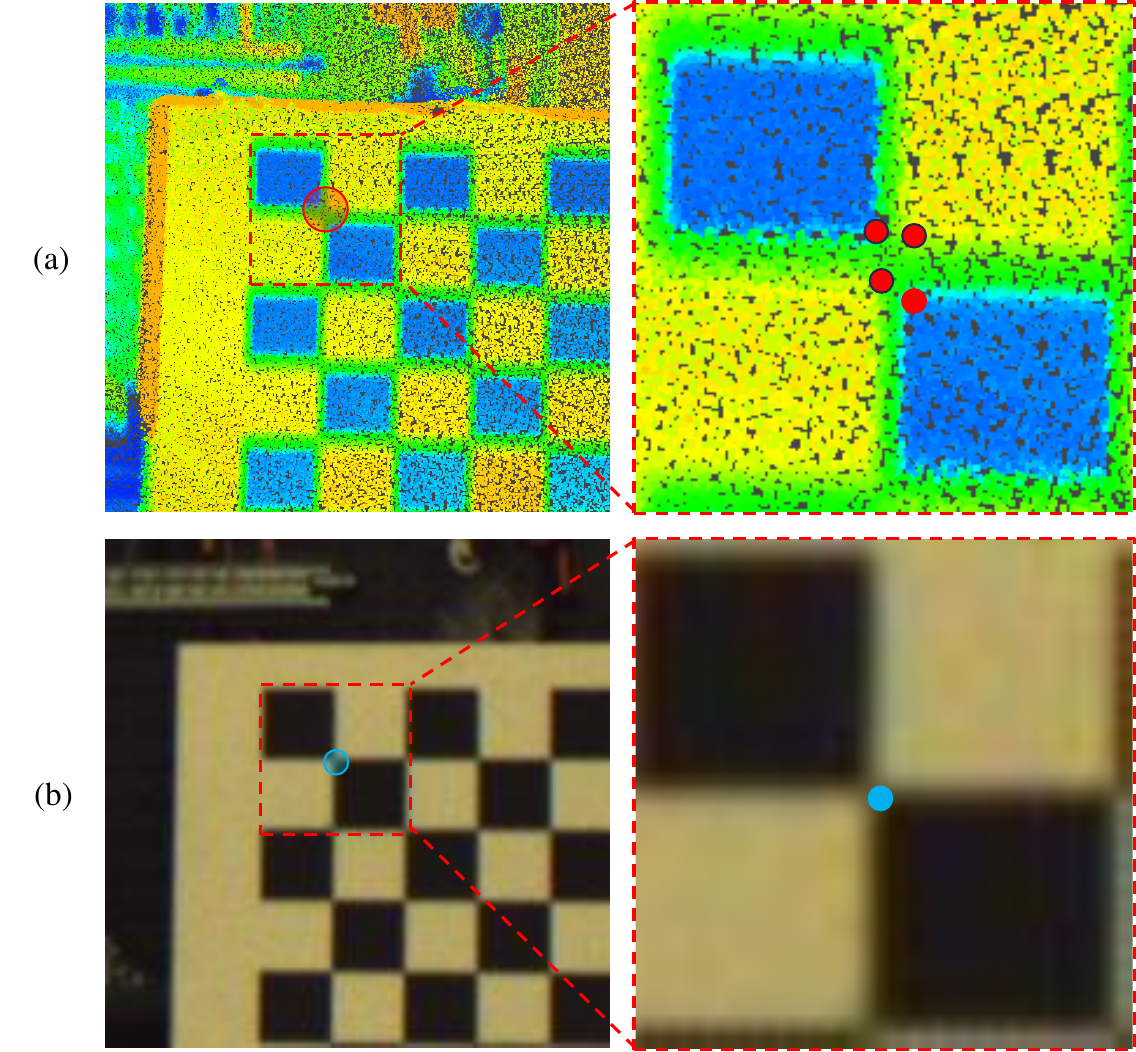}
    \caption{Comparison of corner point detection between LiDAR point cloud and RGB image: (a) possible detection results in the LiDAR point cloud; (b) the detection result in the RGB image. The blue point in (b) has four possible matches in (a).}
    \label{fig.inherent_error}
\end{figure}

\begin{figure*}[t!]
    \centering
    \includegraphics[width=0.999\textwidth]{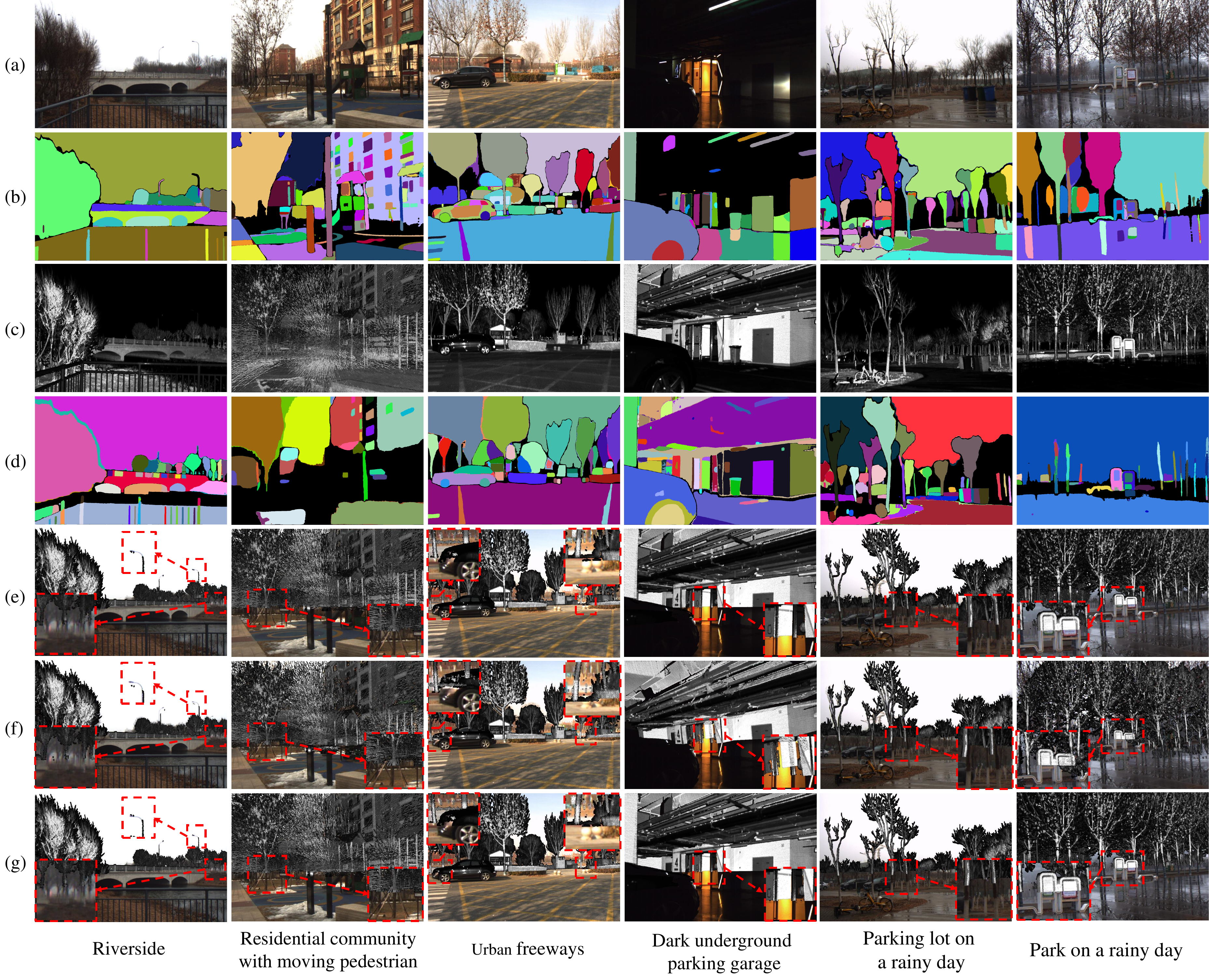}
    \caption{Qualitative comparisons with SoTA target-free LCEC approaches on the MIAS-LCEC-TF70 dataset: (a)-(b) RGB images and their segmentation results; (c)-(d) LIP images and their segmentation results; (e)-(g) experimental results achieved using MIAS-LCEC (ours), HKU-Mars, and DVL, shown by merging LIP and RGB images, where significantly improved regions are shown with red dashed boxes.
    }
    \label{fig.SoTAVisualization}
\end{figure*}

\subsection{Evaluation Metrics}
\label{sec.metrics}

In our experiments, the Euler angle error, with the following expression:
\begin{equation}
    e_r =  \left\|\boldsymbol{r}^* - \boldsymbol{r}\right\|_2,
\end{equation}
where ${\boldsymbol{r}^{*}}$ and ${\boldsymbol{r}}$ represent the estimated and ground-truth Euler angle vectors, computed from the rotation matrices ${^{C}_{L}\boldsymbol{R}}^*$ and ${^{C}_{L}\boldsymbol{R}}$, respectively, and the translation error, with the following expression\footnote{The translation from LiDAR pose to camera pose is $-{{^{C}_{L}\boldsymbol{R}}^{-1}}\boldsymbol{t}$ when (\ref{eq.lidar_to_camera_point}) is used to depict the point translation.}:
\begin{equation}
    e_t= \left\|-{({^{C}_{L}\boldsymbol{R}^*})^{-1}}\boldsymbol{t}^* +{{^{C}_{L}\boldsymbol{R}}^{-1}}\boldsymbol{t}\right\|_2,
\end{equation}
where $\boldsymbol{t}^*$ and $\boldsymbol{t}$ denote the estimated and ground-truth translation vectors, respectively, are used to quantify the performance of target-free LCEC approaches. 

Additionally, we use the following reprojection error
\begin{equation}
\epsilon = \frac{1}{M}\sum_{i=1}^{M}\left\|\boldsymbol{K}(
{^{C}_{L}{\boldsymbol{R}^*}}
 \boldsymbol{p}_{i}^L + 
{^{C}_{L}\boldsymbol{t}^*}
) - \tilde{\boldsymbol{p}}_{i}\right\|_2.
\end{equation}
between 3D LiDAR points and their corresponding 2D image pixels to quantify the performance of LCEC algorithms when using targets.

As illustrated in Fig. \ref{fig.inherent_error}, we observe that LiDAR scanning near textural and geometric discontinuities typically exhibits inherent errors, resulting in a reprojection error of around one pixel. Therefore, in our experiments, we consider results with a reprojection error of less than two pixels to be satisfactory.

\begin{table}[t!]
\caption{
Quantitative comparison of our proposed MIAS-LCEC approach with other SoTA online, target-free approaches on the MIAS-LCEC-TF360 dataset, where the best results are shown in bold type.}
\centering
\settablefont
\begin{tabular}{c|c|rr}
\toprule
Error & Approach & Indoor & Outdoor\\
\hline
\hline
\multirow{5}*{$e_r$ ($^\circ$)} 
&CRLF \cite{ma2021crlf}  &1.469	&1.402\\
 &UMich \cite{pandey2015automatic} &1.802	&2.698\\
&\makecell{HKU-Mars \cite{yuan2021pixel}}  &96.955	&25.611\\
&\makecell{DVL \cite{koide2023general}} &63.003	&46.623\\
&\textbf{MIAS-LCEC (Ours)} &\textbf{0.963}	&\textbf{0.659}\\
\hline
\multirow{5}*{$e_t$ (m)} 
&CRLF \cite{ma2021crlf} &13.484	&0.139\\
&UMich \cite{pandey2015automatic} &0.200	&0.135\\
&\makecell{HKU-Mars \cite{yuan2021pixel}} &4.382 &9.914\\
&\makecell{DVL \cite{koide2023general}} &0.919	&1.778\\
&\textbf{MIAS-LCEC (Ours)} &\textbf{0.182}	&\textbf{0.114}\\

\bottomrule
\end{tabular}
\label{tab.mid360}
\end{table}

\begin{figure}[t!]
    \centering
    \includegraphics[width=0.48\textwidth]{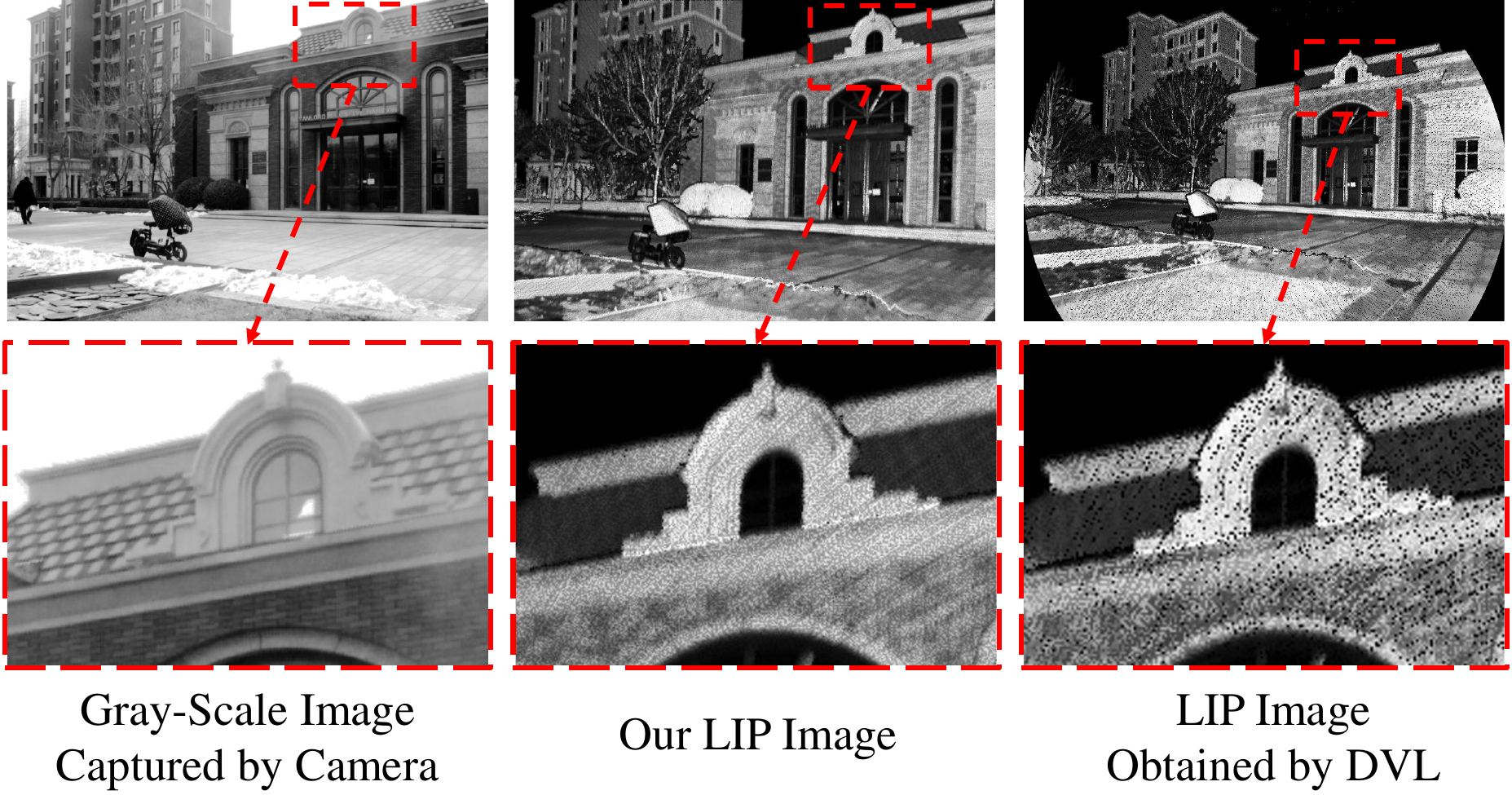}
    \caption{Qualitative comparison between our proposed MIAS-LCEC and DVL in terms of LIP image generation on the MIAS-LCEC-TF70 dataset.}
    \label{fig.LIPimage}
\end{figure}
\begin{figure}[t!]
    \centering
    \includegraphics[width=0.48\textwidth]{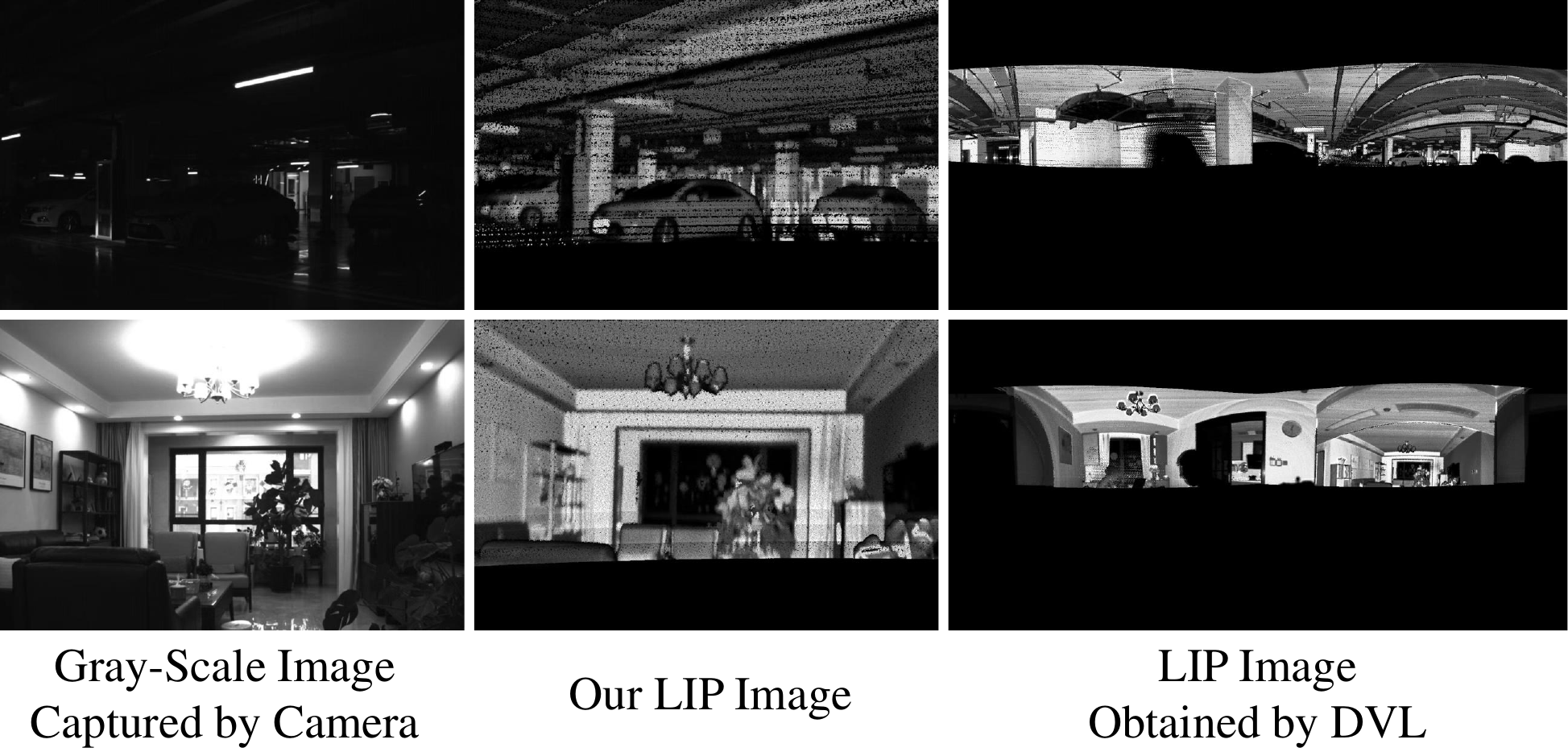}
    \caption{Qualitative comparison between our proposed MIAS-LCEC and DVL in terms of LIP image generation on the MIAS-LCEC-TF360 dataset.}
    \label{fig.mid360}
\end{figure}

\subsection{Comparisons with State-of-the-Art Methods}
\label{sec.comparison_with_sota}

Quantitative comparisons with SoTA approaches on the MIAS-LCEC-TF70 and MIAS-LCEC-TF360 datasets are presented in Tables \ref{tab.ResBySceneCmp} and \ref{tab.mid360}. Additionally, qualitative results for these datasets are illustrated in Figs. \ref{fig.SoTAVisualization}, \ref{fig.LIPimage}, and \ref{fig.mid360}. It is important to note that the results from the first iteration of MIAS-LCEC are reported here because the accuracy achieved is already higher than that of existing SoTA approaches. 

The results shown in Table \ref{tab.ResBySceneCmp} suggest that our method outperforms all other SoTA approaches on a total of 60 scenarios, all captured using a Livox Mid-70 LiDAR. Specifically, MIAS-LCEC reduces $e_r$ by around 30-93\% and decreases $e_t$ by 39-99\%, compared to existing SoTA algorithms. We attribute these performance improvements to the coarse-to-fine correspondence matching pipeline based on LVMs, which sets strict criteria for reliable sparse correspondence selection and propagates these matches to generate dense correspondences, thereby improving the quality of the PnP solutions. It can also be observed that MIAS-LCEC achieves lower mean $e_r$ and $e_t$ values than all other approaches across the total six subsets.
Our method dramatically outperforms CRLF, UMich, and HKU-Mars and is slightly better than DVL in scenarios with low noise and abundant features, while it performs significantly better than all methods in challenging conditions, particularly under poor illumination and adverse weather, or when few geometric features are detectable. This impressive performance can be attributed to MobileSAM, a powerful LVM, capable of learning informative, general-purpose deep features for robust image segmentation. Surprisingly, as observed in Fig. \ref{fig.SoTAVisualization}, MobileSAM can effectively segment both RGB and LIP images captured in challenging conditions, such as dark underground parking garages or during rainy days, where the objects are even unrecognizable to human observers.

Additionally, the results on the MIAS-LCEC-TF360 dataset somewhat exceed our expectations. From Table \ref{tab.mid360}, it is evident that while the other approaches achieve poor performance on this dataset, MIAS-LCEC demonstrates acceptable performance, indicating strong adaptability to more challenging scenarios, with narrow overlapping areas between LIP and RGB images. This performance improvement is primarily attributed to our developed LIP image generation strategy, which incorporates several image pre-processing techniques in our practical implementation to refine the LIP images and align the FoVs between the two sensors as closely as possible. As illustrated in Figs. \ref{fig.LIPimage} and \ref{fig.mid360}, the LIP image generated by DVL contains numerous holes and has a significantly different FoV compared to the RGB image, resulting in unexpected false correspondence matches, which can deteriorate the algorithm's efficiency and robustness. In contrast, MIAS-LCEC can generate LIP images that look as if taken from the same perspective to the actual camera, thus improving the performance of cross-modal mask matching. 

\subsection{Comparison with An Offline, Target-Based Approach}
\label{sec.exp_performance_checkerboards}

\begin{figure}[t!]
    \centering
    \includegraphics[width=0.48\textwidth]{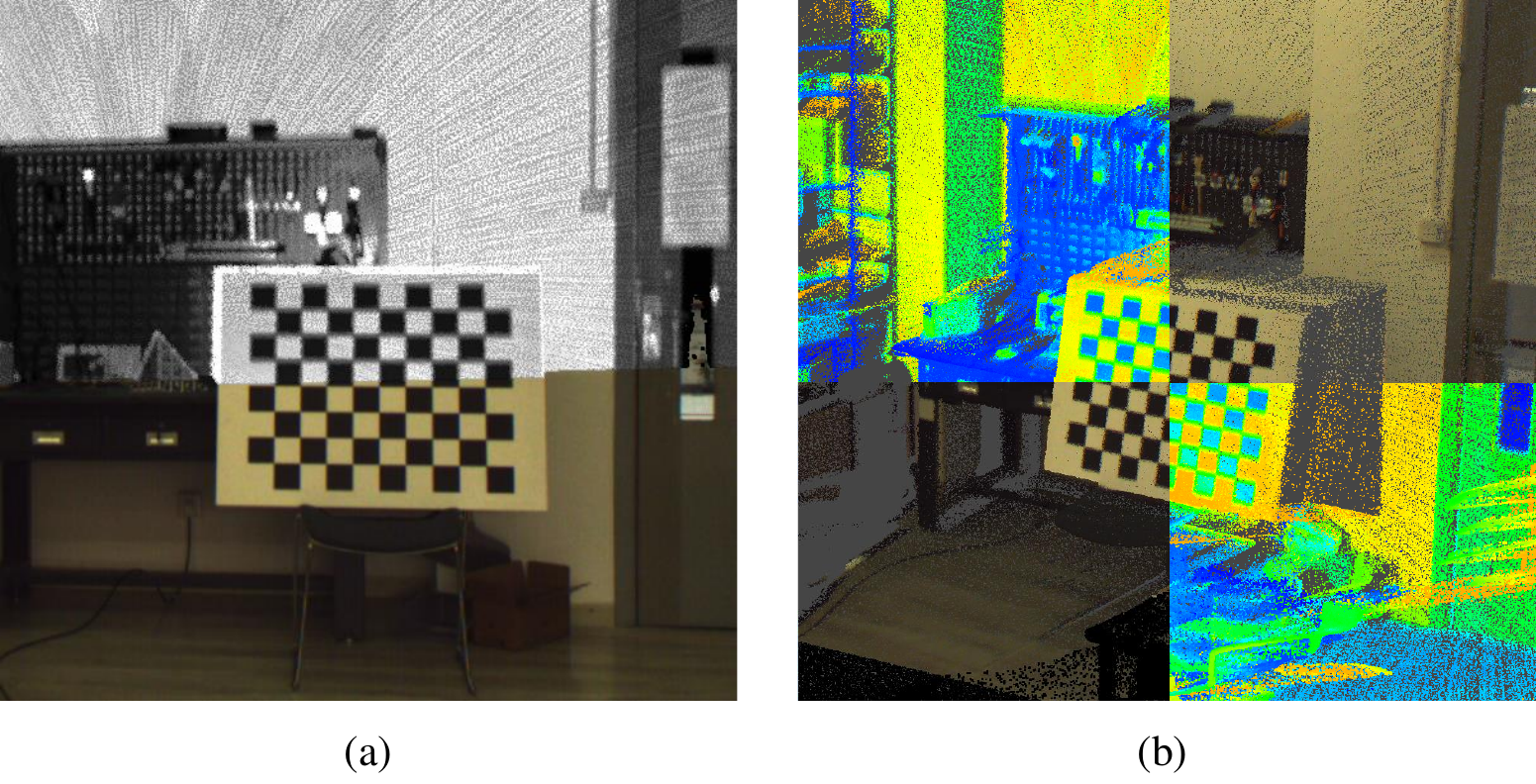}
    \caption{
    Visualization of LCEC calibration results through LiDAR and camera data fusion: (a) fusion of LIP and RGB images; (b) LiDAR point cloud partially rendered by color.}
    \label{fig.rpError_visualize}
\end{figure}

This subsection presents additional experimental results on the MIAS-LCEC-CB70 dataset, comparing our approach with ACSC \cite{cui2020acsc}, a SoTA offline, target-based LCEC algorithm, when calibration targets are available. The checkerboard corner points in both the LIP and RGB images are used as ground-truth correspondences. In contrast to ACSC, which employs ground-truth correspondences to estimate the extrinsic parameters, our approach conducts online, target-free LCEC. The reprojection errors of these correspondences are computed to quantify the performance of both algorithms. As illustrated in Fig. \ref{fig.rpError_visualize}, the visualization of LCEC calibration results through LiDAR and camera data fusion suggests the high accuracy of our approach. Furthermore, as shown in Table \ref{tab.checkerboard}, while MIAS-LCEC achieves satisfactory results, its performance is slightly inferior to that of ACSC. This observation is within our expectations, as ACSC directly minimizes the mean reprojection error of the ground-truth correspondences to determine extrinsic parameters. In contrast, our method relies on distinguishable and matchable masks present in both modalities.

\begin{table}[t!]
\centering
\caption{Comparison of reprojection errors between offline calibration and our proposed MIAS-LCEC approach on the MIAS-LCEC-CB70 dataset.}
\settablefont
\begin{tabular}{c|l|ccc|c}
\toprule
\multirow{2}*{Yaw Angle} &\multirow{2}*{Approach} &\multicolumn{3}{c|}{Distance from Checkerboard} & \multirow{2}*{\makecell{Average $\epsilon$\\
(pixel)}} \\
\cline{3-5}
& &3 m & 4 m &5 m &\\
\hline\hline
\multirow{2}*{$+30^\circ$} 
&Offline \cite{cui2020acsc} &1.451	&1.359	&1.307	&1.372

\\
&\textbf{MIAS-LCEC (Ours)} &1.445	&1.626	&1.653	&1.575

 \\
\hline

\multirow{2}*{$+15^\circ$} 
&Offline \cite{cui2020acsc} &1.589	&1.544	&1.361	&1.498

\\
&\textbf{MIAS-LCEC (Ours)} &1.919	&1.473	&1.386	&1.593

 \\
\hline

\multirow{2}*{$0^\circ$} 
&Offline \cite{cui2020acsc} &1.664	&2.132	&1.329	&1.708

\\
&\textbf{MIAS-LCEC (Ours)} &1.593	&2.086	&1.608	&1.762

 \\
\hline

\multirow{2}*{$-15^\circ$} 
&Offline \cite{cui2020acsc} &1.539	&1.838	&1.743	&1.706
 \\
&\textbf{MIAS-LCEC (Ours)} &1.572	&1.802	&2.204	&1.859\\
\hline

\multirow{2}*{$-30^\circ$} 
&Offline \cite{cui2020acsc} &1.439	&1.534	&1.470	&1.481\\
&\textbf{MIAS-LCEC (Ours)} &1.569	&2.020	&1.883	&1.824\\
\bottomrule
\end{tabular}
\label{tab.checkerboard}
\end{table}

\subsection{LCEC Performance  with Increasing Iterations}
\label{sec.exp_LIP}

\begin{figure}
    \centering
    \includegraphics[width=0.47\textwidth]{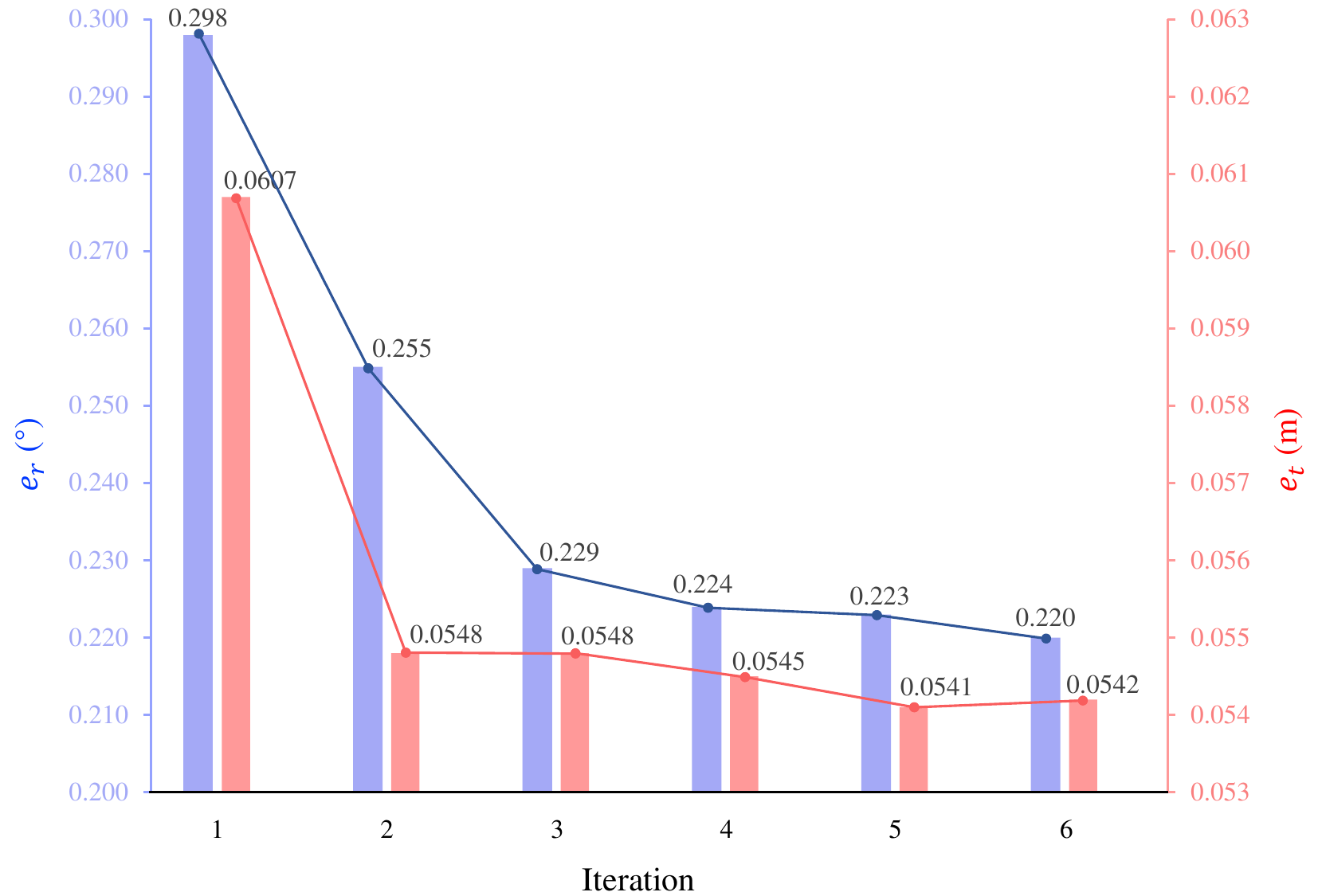}
    \caption{Performance of MIAS-LCEC with increasing iterations on the MIAS-LCEC-TF70 dataset.}
    \label{fig.bar_iter}
\end{figure}

Fig. \ref{fig.bar_iter} illustrates the accuracy of our algorithm with respect to an increasing number of iterations. It is evident that (1) after the first iteration, our approach attains satisfactory accuracy, and (2) after the third iteration, its performance stabilizes and remains relatively consistent (the values of $e_r$ and $e_t$ decrease by approximately 26\% and 11\%, respectively, from the first to the sixth iterations). Therefore, we contend that a single iteration suffices for our MIAS-LCEC approach, striking a balance between accuracy and efficiency. However, additional iterations can certainly be considered when computational resources are abundant.

\section{Conclusion}
\label{sec.conclusion}
This article introduced MIAS-LCEC, a fully online, target-free LiDAR-camera extrinsic calibration approach, developed based on a state-of-the-art large vision model. Compared to prior arts, our approach is more capable of matching cross-modal features and outperforms existing state-of-the-art algorithms. To benefit the robotic community, we also designed a calibration toolbox with an interactive visualization interface based on our developed approach. Extensive experiments were conducted on three real-world datasets to comprehensively evaluate the performance of MIAS-LCEC. The experimental results demonstrate that (1) MIAS-LCEC achieves robust and accurate LiDAR-camera extrinsic calibration without the need for any targets, (2) it demonstrates high adaptability to diverse challenging scenarios by introducing a virtual camera with iterative pose updates to generate more accurate LiDAR intensity projections, and (3) the SoTA image segmentation LVM is successfully applied for this specific task by detecting distinguishable and matchable masks across different modalities. While achieving high accuracy and robustness, the real-time performance of our algorithm still requires improvement, a task we will address in future work.

\normalem

\bibliographystyle{IEEEtran}


\end{document}

%% file: packages.tex
\usepackage{scalerel}
\usepackage{tikz}
\usetikzlibrary{svg.path}
\definecolor{orcidlogocol}{HTML}{A6CE39}
\tikzset{
  orcidlogo/.pic={
    \fill[orcidlogocol] svg{M256,128c0,70.7-57.3,128-128,128C57.3,256,0,198.7,0,128C0,57.3,57.3,0,128,0C198.7,0,256,57.3,256,128z};
    \fill[white] svg{M86.3,186.2H70.9V79.1h15.4v48.4V186.2z}
                 svg{M108.9,79.1h41.6c39.6,0,57,28.3,57,53.6c0,27.5-21.5,53.6-56.8,53.6h-41.8V79.1z M124.3,172.4h24.5c34.9,0,42.9-26.5,42.9-39.7c0-21.5-13.7-39.7-43.7-39.7h-23.7V172.4z}
                 svg{M88.7,56.8c0,5.5-4.5,10.1-10.1,10.1c-5.6,0-10.1-4.6-10.1-10.1c0-5.6,4.5-10.1,10.1-10.1C84.2,46.7,88.7,51.3,88.7,56.8z};
  }
}
\newcommand\orcidicon[1]{\href{https://orcid.org/#1}{\mbox{\scalerel*{
\begin{tikzpicture}[yscale=-1,transform shape]
\pic{orcidlogo};
\end{tikzpicture}
}{|}}}}
\usepackage{hyperref}
\hypersetup{
    colorlinks=true,
    linkcolor=blue,
    filecolor=black,      
    urlcolor=black,
    citecolor=purple
}
\usepackage{tikz-network}